\documentclass[runningheads]{llncs}

\usepackage{eccv}

\usepackage{eccvabbrv}
\usepackage{wrapfig,lipsum,booktabs}
\usepackage{amsmath}
\usepackage{multirow}
\usepackage{pifont}
\usepackage{breqn}
\usepackage{xcolor}
\usepackage{adjustbox}
\usepackage{float}
\usepackage{wrapfig}

\usepackage{graphicx}
\usepackage{booktabs}
\usepackage{comment}
\usepackage{rotating}
\usepackage[accsupp]{axessibility}  
\newcommand{\cmark}{\textcolor{black}{\ding{51}}}%
\newcommand{\xmark}{\textcolor{black}{\ding{55}}}%

\usepackage{hyperref}

\usepackage{orcidlink}
\usepackage{tikz}

\title{Lego: Learning to Disentangle and Invert Personalized Concepts Beyond Object Appearance in Text-to-Image Diffusion Models}
\titlerunning{Lego}

\author{Saman Motamed\inst{1}\orcidlink{0000-0001-5383-5958} \and
Danda Pani Paudel\inst{1}\orcidlink{0000-0002-1739-1867} \and
Luc Van Gool\inst{1,2}\orcidlink{0000-0002-3445-5711}}
\institute{INSAIT, Sofia University, Bulgaria \and
ETH Zurich, Switzerland
}
\begin{document}
\maketitle
\begin{center}
    \centering  
    \captionsetup{type=figure}
    \includegraphics[width=\linewidth]{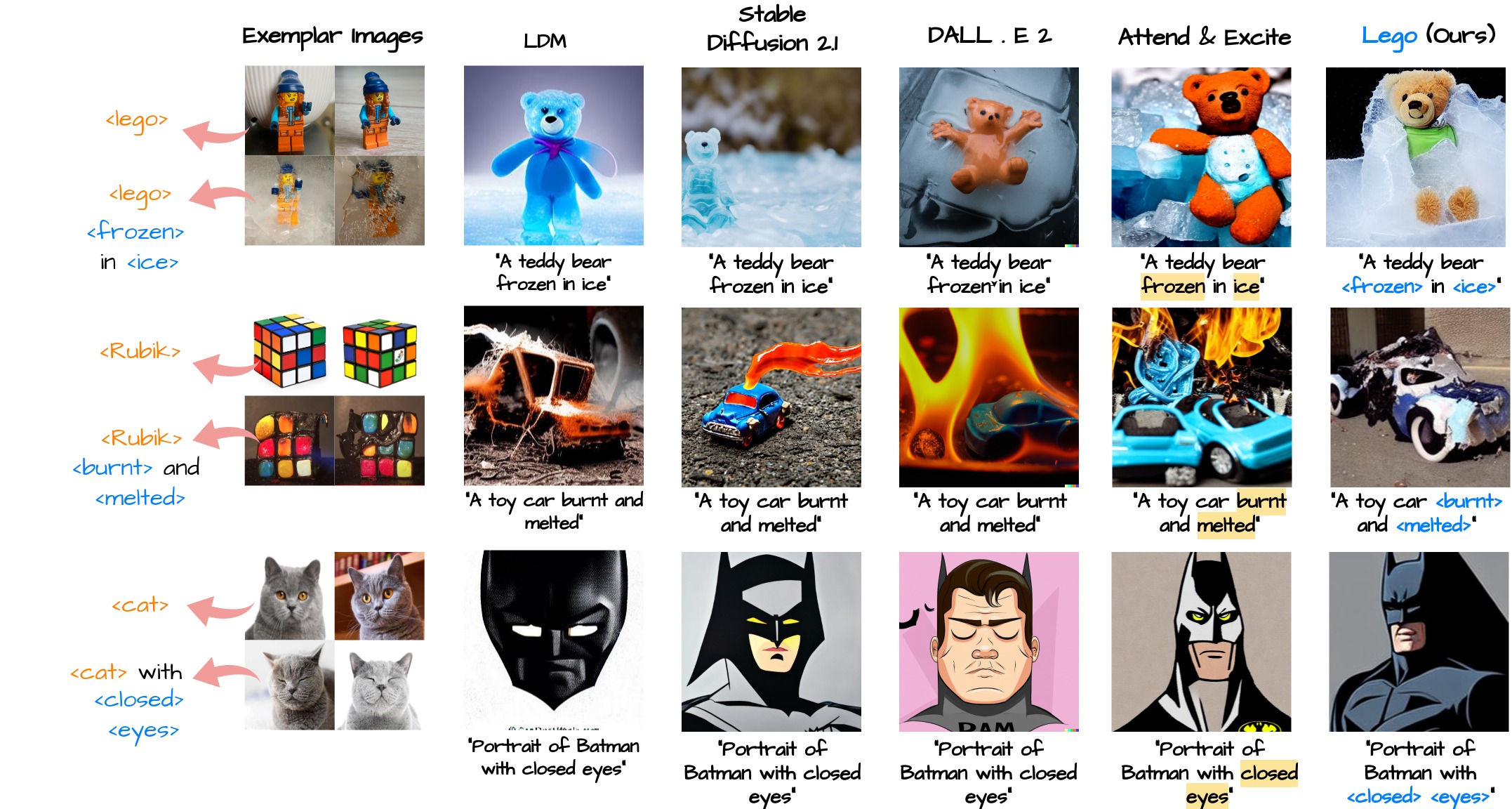}
    \captionof{figure}{\label{fig:first} We showcase Lego's ability to invert concepts of \textit{``frozen in ice''}, \textit{``burnt and melted''}, and \textit{``closed eyes''} using as few as just four example images (two with and two without the concept). Our results cover text-to-image models, including LDM, Stable Diffusion 2.1, Attend and Excite, and closed-source DALL.E 2. Notably, Lego faithfully represents \underline{\emph{intended personalized concepts}}, even with a less capable backbone (LDM), while more powerful models such as DALL.E, though artistically impressive, do not consistently capture the same.}
\end{center}%

\begin{abstract}
 Text-to-Image (T2I) models excel at synthesizing concepts such as nouns, appearances, and styles. To enable customized content creation based on a few example images of a concept, methods such as Textual Inversion and DreamBooth invert the desired concept and enable synthesizing it in new scenes. However, inverting personalized\footnote{Please refer to Figure \ref{fig:td} for our definition of 
\textit{\textbf{personalized}}.}  concepts that go beyond object appearance and style (adjectives and verbs) through natural language, remains a challenge. Two key characteristics of these concepts contribute to the limitations of current inversion methods. \textbf{1)} Adjectives and verbs are entangled with nouns (subject) and can hinder appearance-based inversion methods, where the subject appearance leaks into the concept embedding and \textbf{2)} describing such concepts often extends beyond single word embeddings.

In this study, we introduce Lego, a textual inversion method designed to invert subject entangled concepts from a few example images. Lego disentangles concepts from their associated subjects using a simple yet effective \textbf{Subject Separation} step and employs a \textbf{Context Loss} that guides the inversion of single/multi-embedding concepts. In a thorough user study, Lego-generated concepts were preferred over \textbf{70\%} of the time when compared to the baseline in terms of authentically generating concepts according to a reference. Additionally, visual question answering using an LLM suggested Lego-generated concepts are better aligned with the text description of the concept.
\end{abstract}

\section{Introduction}
\label{sec:intro}
If you saw a Lego figurine frozen in a block of ice or a Rubik's cube melt and deform, how confident would you be in your ability to describe the fine details of the scene by using natural language descriptions alone? And even then, can text-to-image models generate images that accurately follow such text descriptions?
(see Figure \ref{fig:first}). As Mark Twain said; ``Actions speak louder than words'' and describing the fine details of any scene is often more difficult than showing someone / something an example of a similar scene ~\cite{dreambooth, custom-diff, huang2023reversion, epstein2024diffusion, alaluf2023neural, avrahami2023chosen}.

Recently, large text-to-image Diffusion models \cite{ldm,glide,saharia2022photorealistic,yu2022scaling, balaji2022ediffi} have shown promising results in synthesizing high quality images. These models empower users by enabling scene synthesis through natural language descriptions. The ability to craft personalized content with these models, e.g. a scene featuring one's pet dog as Superman, has spurred a research direction aimed at enhancing the user's control for customized content creation \cite{t2i, dreambooth, custom-diff, huang2023reversion, zhang2023adding, hertz2022prompt, sohn2023styledrop, meng2021sdedit, kim2022diffusionclip, ho2022classifier, mokady2023null, gal2023encoder, wei2023elite}. Using a few example images of a concept, text-based inversion methods identify an embedding within the textual embedding space of a text-to-image model's text-encoder that can synthesize that specific concept. This identified embedding can then be injected into various text descriptions, allowing for the synthesis of the concept in diverse scenes. Inversion methods capable of inverting appearance-based concepts include Textual Inversion (TI) \cite{t2i}, DreamBooth \cite{dreambooth}, Custom-Diffusion \cite{custom-diff} and ELITE \cite{wei2023elite}. While TI and ELITE kept the diffusion model frozen, DreamBooth and Custom-Diffusion tuned parts of the model on the example images. Taking a different approach from those works, ReVersion \cite{huang2023reversion} inverts relations between subjects (e.g, under, in, etc.) from a few example images rather than concepts related to subject appearance. 

\begin{figure}
\centering
\scalebox{1}{\includegraphics[width=\linewidth]{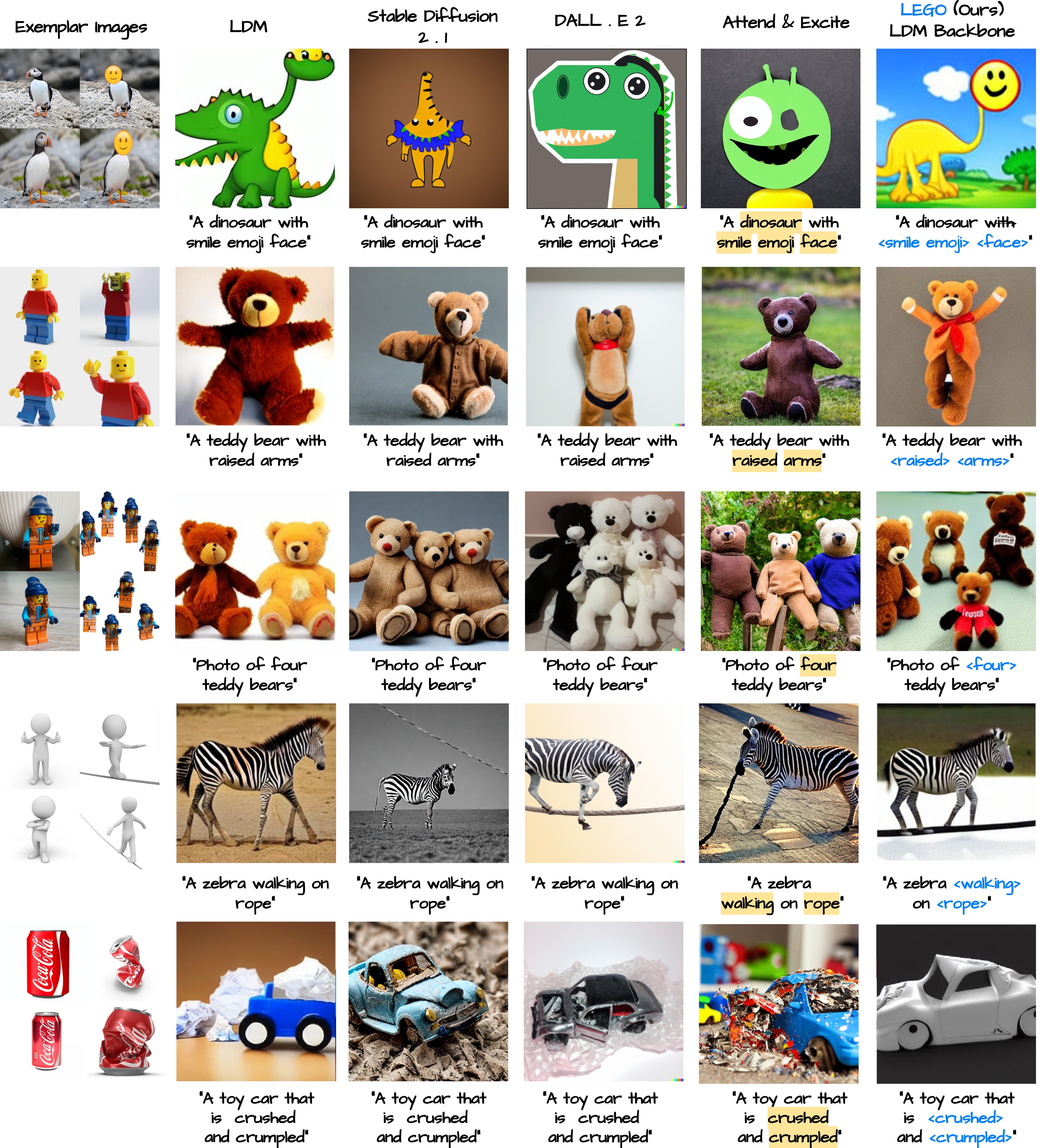}}
\caption{A) We showcase our definition for \underline{\emph{personalized concept}} inversion. While SD 2.1 and DALL.E 2 and 3 create their version of a ``frozen Lego horse in ice'', we are are not only interested in synthesizing the concept, but also doing so such that it follows the example concept of the reference image (personalized) where the concept has unique characteristics ( e.g. cracks and trapped bubbles in the ice). B) We visualize 4 concepts when using LDM with text description of the concept (bottom row) compared to visualizing the concepts after performing Lego inversion using reference images (visualized at the bottom of each Lego generated image) of the concept (top row).}
\label{fig:td}
\end{figure}

\par With vision-language models having shown strong bias towards \textit{nouns} / \textit{objects} \cite{momeni2023verbs, park2022exposing, hendricks2021probing, yuksekgonul2022and} and existing inversion and personalization techniques being predominantly centered around learning appearances, relations and styles, we redirect our attention to object agnostic concepts, specifically, adjectives and verbs. Thus, this paper takes a comprehensive approach to examine the capabilities of text-to-image models in handling adjectives and verbs (see Figure \ref{fig:td}) and the ability of inversion methods in learning to synthesize such concepts. We show that current inversion methods often fail to invert such concepts and our experiments suggest this challenge is due to two key characteristics inherent to these concepts. First, such concepts are entangled with a subject (noun). For instance, the concept of \textit{melting} gives different shapes and characteristics to different subjects it is applied to and current inversion methods are not able to handle subject entangled concepts. Second, describing such concepts frequently extends beyond single word embeddings. For instance \textit{being frozen in ice} is expressed using multiple word embeddings whereas appearance based concepts can have a single word embedding  (e.g., some toy, a pet, etc.).
We introduce \textbf{Lego}, a textual inversion method that augments the TI framework \cite{t2i} with two additional components; a \textbf{\textit{Subject Separation}} step that disentangles a concept from its associated subject by recovering an explicit embedding for the subject and a contrastive \textbf{\textit{Context Loss}} that helps guide multiple embeddings in the textual embedding space, increasing editability and accuracy of the learned embeddings. We show Lego's capabilities to invert various concepts with comparisons to state of the art T2I and inversion methods, and demonstrate that Lego is a reliable and stand-alone inversion method for personalized concepts applicable to any text-conditioned diffusion model.

Our major contributions are summarized below:
 
\begin{itemize}
\par\item We study a new problem, \textbf{\textit{Personalized Concept Inversion}} of adjectives and verbs. We show text-guided image synthesis models and current text based inversion and personalization methods are unable to effectively synthesize such concepts according to a given image of the concept. 
\par\item We propose two modifications to TI; \textbf{Subject Separation} and \textbf{Context Loss} that allow our method to disentangle concepts from subjects and guide the concept's embeddings in the textual embedding space. These modifications together lead to faithful inversion of concepts.
\end{itemize}

\begin{table}[ht]
\caption{A characteristics overview of some recent T2I personalization methods (in order from left to right; Lego, Textual Inversion, ReVersion, Custom-Diffusion, ELITE, ControlNet and Attend and Excite), with a reference to sample images of each method when used for inverting/synthesizing adjective and verb concepts.}
\small 
\centering
\renewcommand{\arraystretch}{0.6} 
\begin{adjustbox}{width=\textwidth}
\begin{tabular}{cccccccc}
\toprule
  & Lego & TI \cite{t2i} & RV \cite{huang2023reversion} & C-Diff \cite{custom-diff} & ELITE \cite{wei2023elite} & CNet \cite{zhang2023adding} & A+E \cite{attend-and-excite} \\
\midrule  
Frozen Model & \cmark & \cmark & \cmark & \xmark & \cmark & \cmark & \cmark \\
\midrule
Text Embedding Optimization & \cmark & \cmark & \cmark & \cmark & \xmark & \xmark & \xmark\\
\midrule
Subject/Concept Disentangling & \cmark & \xmark & \xmark & \xmark & \xmark & \xmark & \xmark \\
\midrule
Multi-Embedding Steering & \cmark & \xmark & \xmark & \xmark & \xmark & \xmark & \xmark\\
\midrule
Sample Results & Figs. \ref{fig:first}, \ref{fig:main-results}, \ref{fig:additional-results} & \multicolumn{2}{c}{Fig. \ref{fig:ablation} - Sec. \ref{supp:ablation}} & Fig. \ref{fig:custom-diff} & \multicolumn{2}{c}{Fig. \ref{fig:cnet-elite}} & Figs. \ref{fig:first}, \ref{fig:main-results} \\
\bottomrule
\end{tabular}
\end{adjustbox}
\label{tab:overview}
\end{table}
\section{Related Work}
\label{sec:related-work}
\noindent\textbf{Diffusion Models} \cite{ddpm, gu2022vector, song2020score,song2020denoising, sohl2015deep}  have become the go-to generative model over their counterparts \cite{brock2018large, karras2021alias, goodfellow2020generative, ding2021cogview, chang2023muse, gafni2022make} with their superior synthesis quality and more stable training. Recently, text-to-image (T2I) diffusion models \cite{nichol2021glide, ramesh2021zero, ramesh2022hierarchical, rombach2022high, saharia2022photorealistic} have shown promise in enabling an intuitive interface for users to control image generation, using natural language descriptions. However, gaining granular control and customized content generation has proven difficult with natural language descriptions alone \cite{feng2022training, liu2022compositional, Wu_2023_CVPR,lee2023aligning}. Addressing this difficulty has started a line of research for inverting desired concepts in these large models and better tuning them for customized content creation. In Table~\ref{tab:overview} we describe the characteristics of the most relevant personalization models and give further detail on some of the recent works that attempt to solve this problem below.

\begin{figure}
\centering
\scalebox{1}{\includegraphics[width=0.7\linewidth]{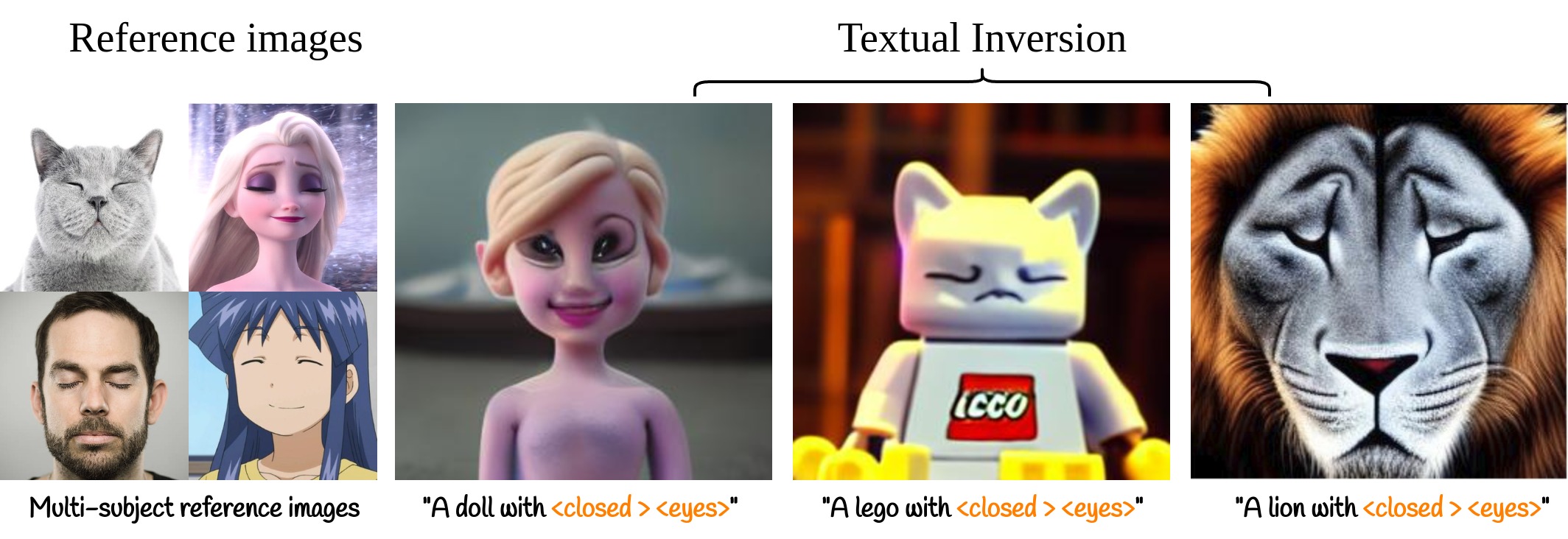}}
\caption{Textual Inversion is not able to learn the concept of ``closed eyes'' from multiple subjects without the appearance of the sample subjects leaking into the concept embedding.}
\label{fig:tifails}
\end{figure}

\noindent\textbf{Textual Inversion.} Given a T2I model, Textual Inversion \cite{t2i} is tasked with finding a pseudo-word's embedding that can represent a subject, given as few as 3-4 exemplar images of that subject. Without fine-tuning any part of the network, TI searches the textual embedding space of the diffusion model's vision-language encoder (BERT \cite{devlin2018bert}, CLIP \cite{radford2021learning}, etc.) to find an embedding that can synthesize the given object in the reference images. In Figures  \ref{fig:tifails} and \ref{fig:ablation} and Supplementary Section \ref{supp:ablation}),  we show that using multiple images of a concept and performing TI is not enough for inverting concepts. TI uses a single embedding that is forced to not only learn to represent the concept, but also the subject appearance, leading to appearance leakage.

\noindent\textbf{DreamBooth + Custom Diffusion.}
With similar objective as TI, DreamBooth~\cite{dreambooth} and Custom-Diff~\cite{custom-diff} not only perform optimization in the textual embedding space, but also train parts of the T2I diffusion network with the exemplar images in order to achieve a better representation of the given concept. DreamBooth and Custom-Diff achieve better concept representation compared to TI. Custom-Diff is similar to DreamBooth with a few differences; 1) allowing multiple concepts being learned simultaneously, 2) light weight tuning by only updating the cross-attention parameters of the diffusion network and 3) a regularization step to stop language drift during tuning. Section \ref{sec:mcdiff-results} shows how these models fail to invert subject entangled concepts.

\noindent\textbf{Attend-and-Excite.} Feeding long text descriptions to T2I models often leads to catastrophic forgetting; where some words in the sentence fail to appear in the generated image. Attend and Excite \cite{attend-and-excite} enforces the cross-attention units of the diffusion network to be activated for the user-selected tokens, encouraging the model to generate all subjects described in the text prompt. In Figures \ref{fig:first} and \ref{fig:main-results}, we show that selecting the concept tokens and using the Attend and Excite method is not enough to generate the concepts. 

\noindent\textbf{ReVersion.} Relation Inversion \cite{huang2023reversion} is the only work besides ours that focuses on inverting non-appearance concepts, namely relations between subjects. ReVersion uses a contrastive loss based on InfoNCE \cite{oord2018representation} that steers the relation embedding towards the embeddings of prepositions (preposition prior), with the observation that in natural language, prepositions express the relation between subjects. ReVersion also uses natural language descriptions of the subjects in the exemplar images to separate the subjects from the relation embedding. Relations control the positioning of subjects with respect to one another and hence, the subjects stand alone. This facilitates the use of natural language descriptions alone for separating the relation embedding from subject appearance. In contrast, our focus lies in studying concepts entangled with the subject. Our ablation study (Section \ref{sec:ablation}) shows that ReVersion's framework fails to invert subject-entangled concepts.

\section{The Concept Inversion Problem}
\begin{figure}
\centering
\scalebox{1}{\includegraphics[width=\linewidth]{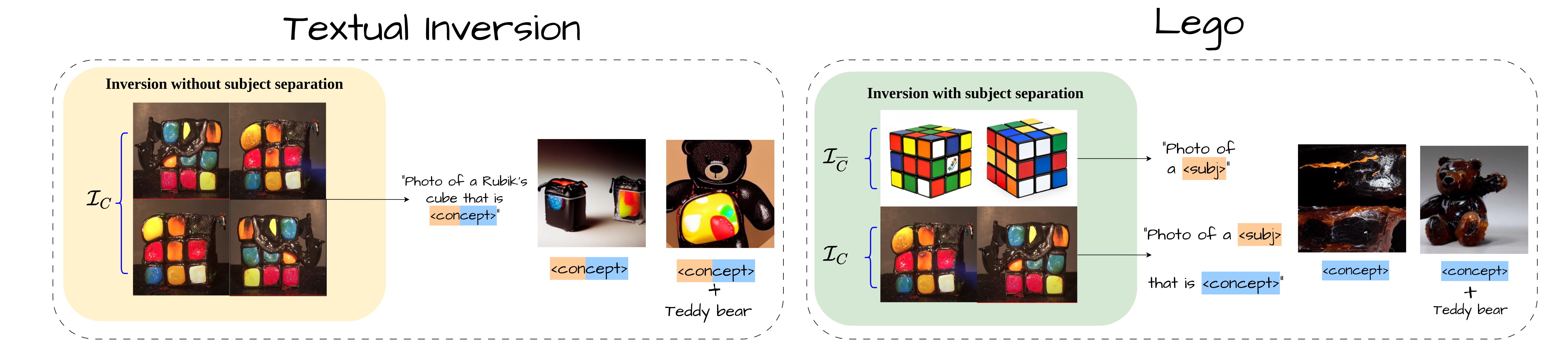}}
\caption{Right figure is an overview of Lego's objective and the Subject Separation step. Learning an explicit embedding to represents the subject (Rubik's cube) allows the concept (``melted'') embedding to dissociate from the subject's appearance features, as visualized by $<$concept$>$ embedding (highlighted in blue). The left figure depicts the framework that uses concept only images (same setting as TI, DreamBooth, etc.). In this setting, the subject's features leak into the $<$concept$>$ embedding, (highlighted in orange and blue), as shown by the concept's visualization which evinces both melting effects and Rubik's cube features. }
\label{fig:prblm-statement}
\end{figure}

            The goal of Lego is to learn embeddings ${\mathcal{CPT} =\{<\!cpt_{1}\!>, ..., <\!cpt_{n}\!>\}}$ that represent a concept $\mathbf{C}$, from a few exemplar images. Let ${\mathcal{I} = \{ I_{1}, I_{2}, ... , I_{m}\}}$ be a small set of such exemplar images involving a common subject $\mathbf{S}_e$. Lego requires
            a clear separation between images of a subset with concepts, say $\mathcal{I}_{C}$, and the same without, say $\mathcal{I}_{\overline{C}}$, such that $\mathcal{I}=\mathcal{I}_{C}\cup\mathcal{I}_{\overline{C}}$.
            For the example shown in Figure~\ref{fig:prblm-statement} (right), the caption of all the images in $\mathcal{I}_{\overline{C}}$ is ``photo of a Rubik's cube'', whereas the same in $\mathcal{I}_C$ is ``photo of a Rubik's cube that is melted'', where the exemplar subject $\mathbf{S}_e$ is ``Rubik's cube'' and the general concept of interest $\mathbf{C}$ is ``melted''. In this setting, we wish to learn the embeddings $\mathcal{CPT}$ corresponding to the concept $\mathbf{C}$ such that the concept can be transferred to any novel target subject $\mathbf{S}_t$-- which is ``a teddy bear'' in the very same example.

Our experiments show the inherent difficulty in synthesizing accurate representations of verb and adjective concepts using T2I models with natural language guidance.  The ability to learn embeddings capable of representing such concepts using as few as four exemplar images allows the user to have greater control over T2I models. In the context of T2I models, adjective and verb concept inversion, where the concept is entangled with subjects, has not been previously explored. We demonstrate that such entanglement poses challenges for inversion methods that have previously aimed at subject/concept separation, as observed in other concept types like relations \cite{huang2023reversion}. 

\begin{figure}[t]
\centering
\includegraphics[width=\linewidth]{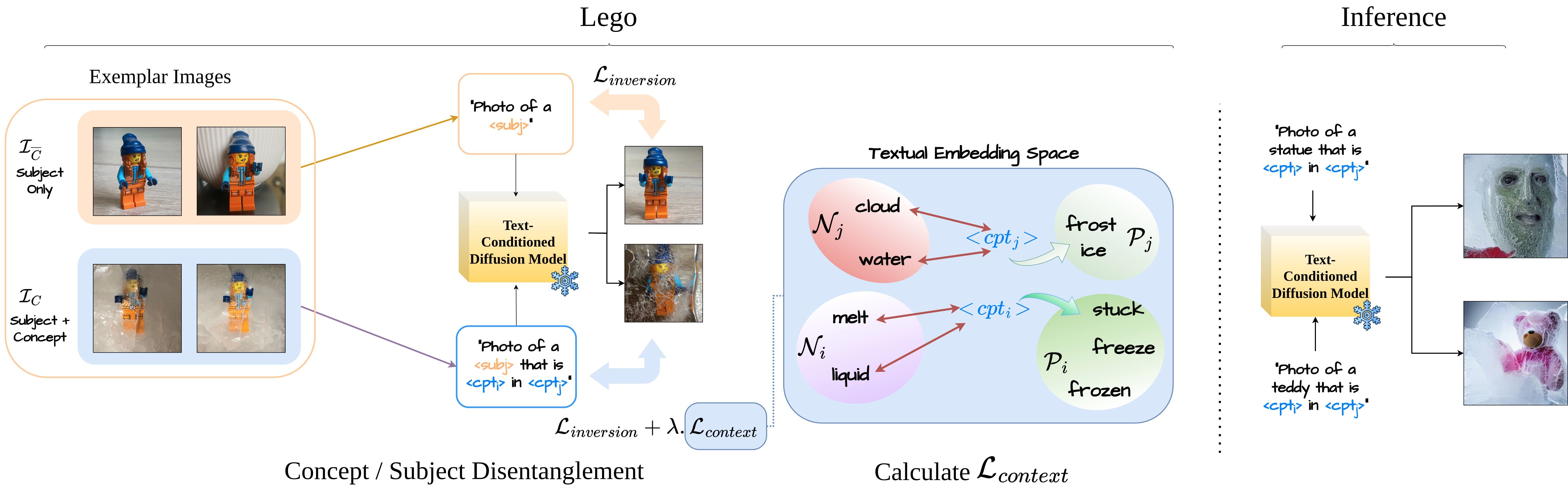}
\caption{An overview of Lego's framework. From left to right, during embedding optimization, Lego dedicates an embedding  \(<\!subj\!>\) for inverting the subject $\mathbf{S}_e$ in the exemplar images $\mathcal{I}_C$  and $\mathcal{I}_{\overline{C}}$. This stops appearance leakage to the concept embeddings. Each concept embedding ( \(<\!cpt_{i/j}\!>\)) is separately steered towards user defined words ($\mathcal{P}_{i/j}$) that correspond to the embedding's semantic word and away from antonyms of those words ($\mathcal{N}_{i/j}$). After the inversion, the learned embeddings can together be applied to different target subjects $\mathbf{S}_t$ (``Statue'' and ``Teddy bear'') to manifest the concept in new scenes.}
\label{fig:method}

\end{figure}

\section{Learning Concepts Beyond Appearance}
\label{sec:method}

\subsection{Preliminaries}
\par\noindent\textbf{T2I Diffusion Models}. \quad
Diffusion models \cite{ddpm, sohl2015deep, dhariwal2021diffusion} are a class of generative models that learn to generate novel scenes by learning to gradually denoise samples from the Gaussian prior $\text{x}_{T}$ (trained by adding noise $\epsilon \sim \mathcal{N} (0,1)$ to $\text{x}_{0}$) back to the image $\text{x}_{0}$. In this work, we study T2I models, namely the Latent Diffusion Model (LDM) \cite{ldm}. Instead of directly adding noise and learning to denoise images, LDM operates on a pretrained autoencoder's projection of images in some latent space. LDM enables T2I generation by conditioning the denoising network $\epsilon_{\theta}(.)$ on the encoding of text descriptions $c$ using a text encoder $\tau_{\theta}(.)$ such as CLIP \cite{radford2021learning} or BERT \cite{devlin2018bert}. To sample images using a trained latent diffusion model $\epsilon_{\theta}(.)$, we iteratively denoise a noise latent $\text{x}_{t}$ for $t$ steps, using the predicted noise  $\epsilon_{\theta}(\text{x}_t,t,\tau_{\theta}(c))$ to get $\text{x}_0$. $\text{x}_0$ is then mapped to the image space using some pre-trained decoder. The LDM loss is: $\mathcal{L}_{LDM}(\theta) := \mathbb{E}_{t, \text{x}_0,\epsilon}[
\|   \epsilon - \epsilon_{\theta} (\text{x}_t, t, \tau_{\theta}(c))
\|^{2}] \label{sa}$.

\par\noindent\textbf{Inversion in the Textual Embedding Space.} \quad Current textual inversion methods focus on either appearance inversion of a subject~\cite{t2i, custom-diff, dreambooth, wei2023elite, Vinker_2023} or inverting a relation between subjects \cite{huang2023reversion}. Given a few exemplar images of a subject or relation, the aim is to find a text embedding $<$$emb^*$$>$ in the output space of $\epsilon_{\theta}(.)$, such that injecting $<$$emb^*$$>$ in any encoded text $\tau_\theta(c)$ allows the reconstruction of that concept in a new context defined by description $c$. The embedding inversion loss is defined as:
\begin{equation}
\mathcal{L}_{\text{inversion}}= \mathbb{E}_{t, \text{x}_0,\epsilon}[
\| \epsilon - \epsilon_{\theta} (\text{x}_t, t, \tau_{\theta}(c)) \|^{2}],
\label{eq:inversion}
\end{equation}
such that: $<\!emb^*\!> = \underset{<emb>}{\mathrm{\text{arg min}}} (\mathcal{L}_{\text{inversion}} )$, where 
$<$$emb$$>$ is the concept embedding being optimized, and is fed into the pretrained T2I model as part of the text description c.
\subsection{Method Overview}
\par In this section, we present \textbf{Lego}, an inversion method for extracting personalized concepts (adjectives and verbs) from exemplar images. Lego augments Textual Inversion with two modifications; \textit{\textbf{Subject Separation}} and \textit{\textbf{Context Loss}}. Lego aims to invert concepts that move beyond objects' relations, styles, and appearances. These concepts are considered to be entangled with the subject of the exemplar images. For instance, The concept of ``melting'' is not standalone and changes the features of the subject it is applied to, whereas in subject inversion (TI \cite{t2i}, Custom-Diff \cite{custom-diff}, etc.), the subject's appearance does not change and stands alone. Similarly, in relation inversion (ReVersion \cite{huang2023reversion}), the relation defines how different subjects interact and does not change the subjects' appearance features. 
\subsection{Subject Separation} \hspace{1em} Appearance inversion focuses on low-level features in order to learn a single style or subject appearance from a few sample images. For such subject-centric inversions, a pixel level reconstruction loss (Equation \ref{eq:inversion}) is often sufficient to find an embedding that represents the subject. Inverting relations between subjects (ReVersion) is a higher-level concept that requires more than a pixel level loss which ReVersion addresses using a preposition prior (Section \ref{sec:related-work}). Relations between subjects however, are not entangled with the subjects' appearance, hence ReVersion is able to detach the relation embedding from the subject embedding by steering it away from the embedding of natural language words that describe the subjects (e.g., while inverting the relation of Batman and Superman sitting back to back, the relation embedding is steered away from embeddings of ``Batman'' and ``Superman'') . Our concepts however are entangled with subjects such that the same approach as ReVersion leads to appearance leakage (see Figures \ref{fig:prblm-statement} - left and \ref{fig:ablation}) and does not allow concepts to be separated from the subjects.

In order to learn embeddings $\mathcal{CPT}$ representing the concept $\mathbf{C}$ disentangled from the exemplar subject $\mathbf{S}_e$, we dedicate an additional embedding $<\!subj\!>$ to separately represent the subject $\mathbf{S}_e$. This embedding gets optimized twice using (i) $\mathcal{I}_{\overline{C}}$ and (ii) $\mathcal{I}_{{C}}$, separately. 
 Learning $<$\textit{subj}$>$ from subject-only images $\mathcal{I}_{\overline{C}}$ naturally enables the concept embeddings $\mathcal{CPT}$ to not have to learn the subject's appearance, while reconstructing $\mathcal{I}_{{C}}$ using both $<$\textit{subj}$>$ and $\mathcal{CPT}$  embeddings (see Figure~\ref{fig:prblm-statement}). In other words, learning the subject embedding without concept and using its embedding to generate the subject with concept during optimization allows us to better perform the desired disentanglement.  Our experiments illustrate that the proposed learning setup prevents the subject-specific features to leak into the concept embedding (see Figure \ref{fig:td} - B). In Figures \ref{fig:prblm-statement} and \ref{fig:ablation} and in the Supplementary Section~\ref{supp:ablation}, we show that the absence of the proposed Subject Separation leads to undesired outcomes. For instance, Figure \ref{fig:prblm-statement} (left) shows how not performing Subject Separation in learning the concept of \textit{melting} from Rubik's cube images leads to the ``$<$\textit{melted}$>$ toy bear'' to have Rubik's features and learning the concept of \textit{closed eyes} from multiple subjects in Figure \ref{fig:tifails} leads to the appearance of those subjects leaking into new subjects. In contrast, learning an explicit embedding for the cube - Figure~\ref{fig:prblm-statement} (right) - keeps the Rubik's features disentangled from the concept embedding.
 
\subsection{Contrastive Context Guidance} 
\hspace{1em} Concepts of our interest (e.g., frozen in a block of ice) often require a multi-word description. Hence in this work, we allow for learning multiple word embeddings per concept. More specifically, Lego enables learning multiple embeddings (${\mathcal{CPT} = \{ <\!cpt_{1}\!>, ..., <\!cpt_{n}\!>\}}$) for a single concept -- such that combining all $n$ embeddings represents the concept,-- unlike, TI \cite{t2i} and Custom-Diff \cite{custom-diff} which learn concepts described by a single word embedding. 
Inspired by success of contrastive losses for representation learning \cite{oord2018representation, 2020rep,Yuan_2021_CVPR, huang2023reversion}, we employ an InfoNCE-based \cite{oord2018representation} loss to learn the embeddings $\mathcal{CPT}$ in a contrastive setting.
	
Learning concept embeddings in the contrastive setting however requires positive and negative embedding sets. Let  $\mathcal{P}_i=\{P_{ik}\}$ and $\mathcal{N}_i=\{N_{ik}\}$ respectively be the positive and negative embedding sets corresponding the concept embedding $<\!cpt_{i}\!>$. Note that each embedding $<\!cpt_{i}\!>$ corresponds to a semantic word. We form the sets $\mathcal{P}_i$ and $\mathcal{N}_i$ by embedding the synonyms and antonyms, respectively, of the semantic word corresponding to $<\!cpt_{i}\!>$. Please refer to Section \ref{supp:design} for more detail on choosing $\mathcal{P}_i$ and $\mathcal{N}_i$. In this manner, for the general concept $\mathbf{C}$, we obtain a set of triplets $\{(<\!cpt_i\!>, \mathcal{P}_i, \mathcal{N}_i)\}_{i=1}^n$. This triplets' set is then used to compute our modified InfoNCE loss, referred here as $\mathcal{L}_{\text{context}}$, which is given by,
 
\newcommand{\cpt}{<cpt_{i}>}
\newcommand{\sumCPT}{\sum\limits_{i=1}^{n}}
\newcommand{\sumP}{\sum\limits_{\!k=1\!}^{|\!\mathcal{P}_{i}\!|}}
\newcommand{\sumN}{\sum\limits_{\!k=1\!}^{|\!\mathcal{N}_{i}\!|}}
\newcommand{\expCPTP}{e^{\cpt^\intercal \cdot P_{ik}}}
\newcommand{\expCPTN}{e^{\cpt^\intercal \cdot N_{ik}}}
\begin{equation}
\!\mathcal{L}_{\text{context}}\!=\!-\!\sumCPT\log\!\frac{ \sumP \expCPTP}
{\sumP\!\expCPTP\!+\!\sumN\!\expCPTN}. 
\label{eq:1}
\end{equation}
During the learning process, the context loss guides each concept embedding $<\!cpt_{i}\!>$ individually, towards the embeddings of the respective positive words and away from the negative ones. Figure \ref{fig:method} shows an example where we want to capture the concept of ``frozen in ice'' with two embeddings $<\!cpt_{i}\!>$ and $<\!cpt_{j}\!>$, one representing ``frozen'' and one representing ``ice''. When combined together in a sentence, they express the concept $\mathbf{C}$ in the exemplar images $\mathcal{I}_{C}$. In Section \ref{sec:ablation}, we provide examples of concept inversion with and without the context loss. Our objective of learning descriptive multi-word concepts, enabled by our context loss, requires the embeddings to be steered towards their corresponding semantic word, each in a different part of the text-embedding space. Furthermore, the negative sets of words allow Lego's embeddings to be steered away from words that can disturb the concept inversion. By disturbing the inversion, we refer to words that are in close distance to our concept embedding, yet associated with a concept we do not want to represent.  
For instance, numbers are closely embedded in the text embedding space. In order to accurately invert cardinality, say number $3$, we can construct a negative set of words comprised of $\{1, 2, 4, 5\}$.

\subsection{Lego}
Lego uses the inversion loss in Equation~\ref{eq:inversion} to learn an embedding $<$$subj^*$$>$ that represents the subject of the exemplar images, by optimizing an embedding $<$\textit{subj}$>$. While optimizing the concept embeddings, the weighted sum of the inversion loss and our context loss is used to obtain the concept embeddings $<\!\mathcal{CPT^*}\!>$. Below we show both the subject and concept inversion losses. 
\begin{equation}
\begin{split}
 <\!subj^*\!> &= \underset{<subj>}{\mathrm{\text{arg min}}} (\mathcal{L}_{\text{inversion}} ),  \\
<\!\mathcal{CPT^*}\!> &= \underset{<cpt>}{\mathrm{\text{arg min}}}  (\mathcal{L}_{\text{inversion}} + \lambda . \mathcal{L}_{\text{context}}).
\end{split}
\end{equation}
These two embeddings' recoveries, corresponding to subject and concepts, can be thought as two parallel processes, being optimized at the same time, acting on the same image generator and the set of exemplar images. Such clear separation in learning process enables us to achieve the desired disentanglement.

 \section{Experiments}
\label{sec:experiments}

\subsection{DreamBooth and Custom Diffusion}
\label{sec:mcdiff-results}
\begin{figure}
\centering
    \includegraphics[width=0.7\linewidth]{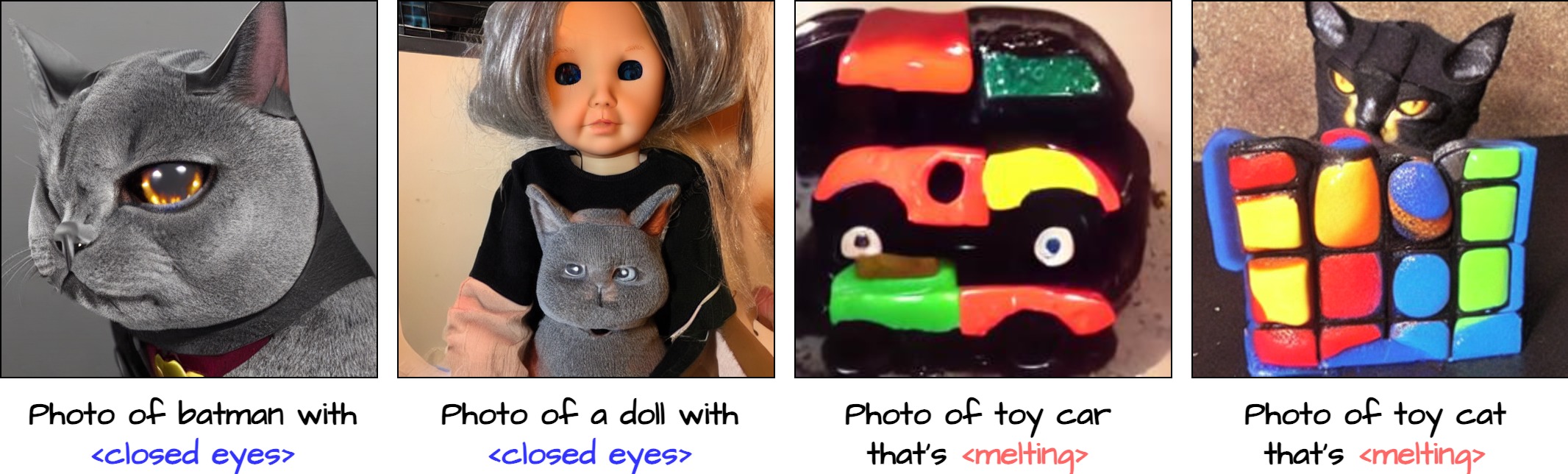}
\caption{\label{fig:custom-diff} Custom-Diff is unable to separate the concept embedding from the subject appearance. Learning ``closed eyes'' from cat images and ``melted'' from Rubik images will carry the subject features when applied to new subjects (appearance leakage).}
\end{figure}

Both DreamBooth \cite{dreambooth} and Custom-Diff \cite{custom-diff} pursue the same objective by tuning parts of the diffusion model, while optimizing a word embedding in order to achieve better accuracy in synthesizing personalized concepts compared to methods like Textual Inversion that keep the model frozen. Since Custom-Diff allows learning multiple embeddings at the same time, we carried out our experiments using Custom-Diff. With the same exemplar images used in Lego, we trained Custom-Diff to learn the embeddings of ``Rubik's cube'' and ``melting'' from Rubik's $\mathcal{I_{C}}$ and $\mathcal{I}_{\overline{C}}$ images and the concept of ``cat'' and ``closed eyes'' from the cat images. Figure \ref{fig:custom-diff} shows that inverting concepts while training the network does not allow for dissociation of concepts from subjects and leads to appearance leakage. We show how Custom Diffusion's learned embedding for ``melting'', when applied to a \textit{toy car} and a \textit{toy cat}, inserts the Rubik's cube features in those images and similarly, the learned concept for ``closed eyes'' from the cat images, inserts the cat's features into images of \textit{Batman} and \textit{a doll}.

\subsection{Lego}
We tested Lego's capabilities in inverting 10 various concepts; from controlling cardinality of subjects (\textbf{3}, \textbf{4}, \textbf{5}) to concepts that deform the subject (\textbf{melting}, \textbf{crumpling}), to the subject performing an action (\textbf{closed eyes}, \textbf{walking on a rope}, \textbf{arms raised}) and change of state and appearance of the subject (\textbf{frozen in ice} and \textbf{having a smiley emoji face}). For each experiment, 4 example images were used (2 with and 2 without the concept). 

\begin{figure}
\centering
    \includegraphics[width=\linewidth]{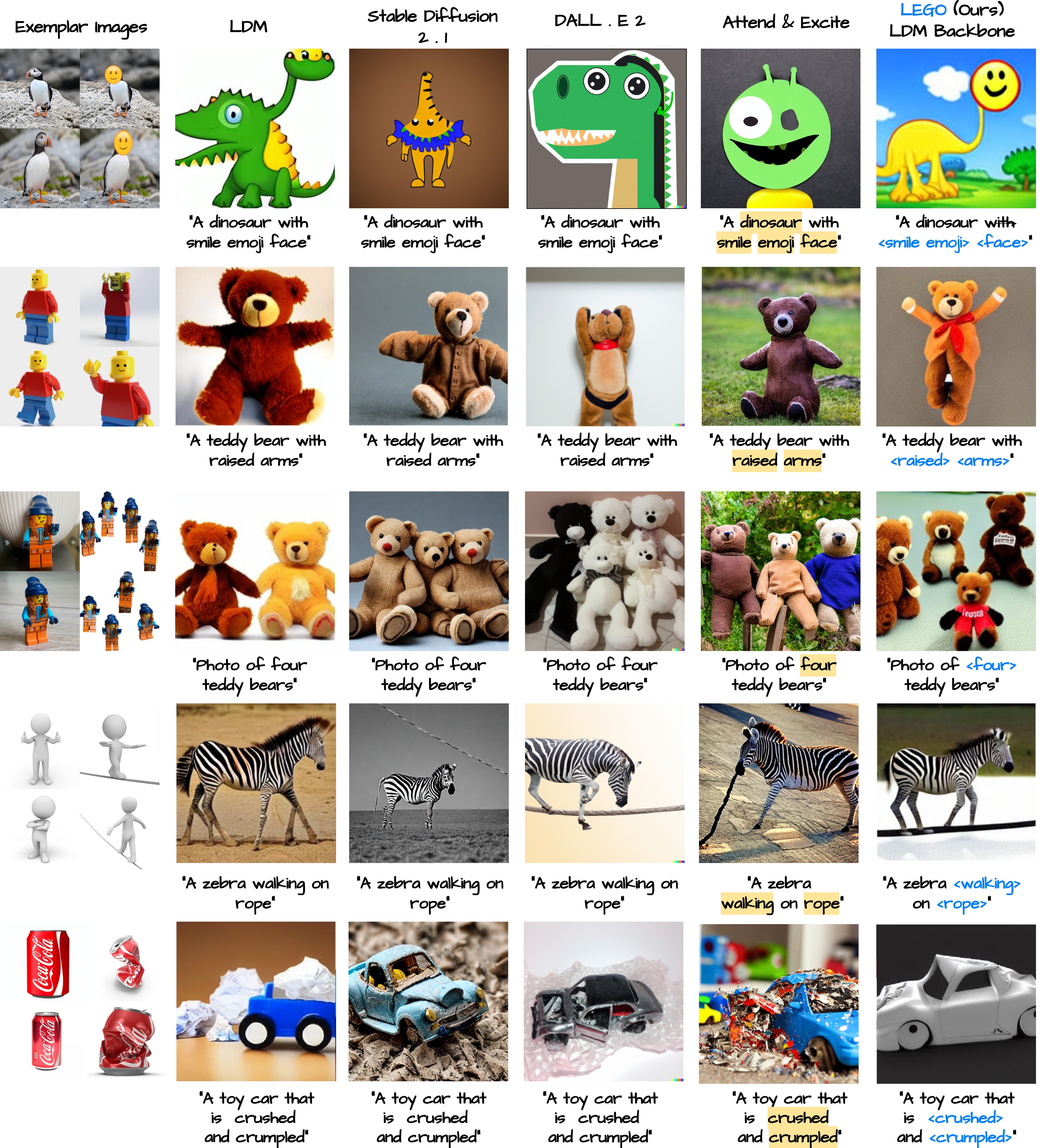}
\caption{\label{fig:main-results} Qualitative comparison of \textbf{Lego} with an LDM backbone and learned concepts from exemplar images, compared to text-guided models such as LDM, SD 2.1, DALL.E 2 and Attend \& Excite (highlighted words are the given token). Zoom in for details.}
\end{figure}

\begin{figure}
\centering
    \includegraphics[width=\linewidth]{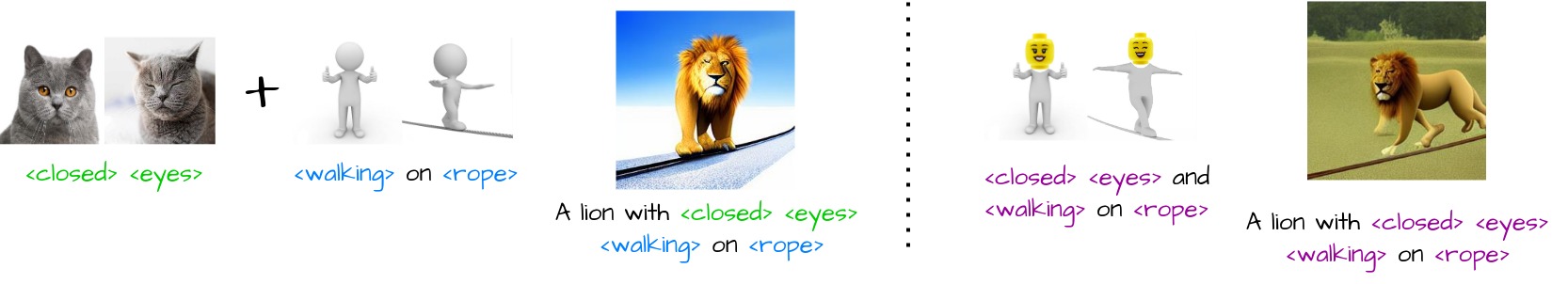}
\caption{\label{fig:composition} This figure (left) shows Lego's composition capability of combining different concepts learned from different images and (right) learning more complex multi-word embedding concepts from a single example.}
\end{figure}

\subsection{Results}
\par\noindent In Figures \ref{fig:first} and \ref{fig:main-results}, we show Lego's results for a subset of the concepts and provide comparisons with the latest SOTA methods for T2I generation. Should certain results be selectively highlighted, it may skew the reader's perception of Lego's effectiveness. We note that Lego can learn to authentically represent the concept of the example images even where more recent models such as Stable Diffusion 2.1 and DALL.E that can generate the concept, are not able to do so faithful to the example image. Similar to TI, Lego is standalone and can be applied to any T2I model. We will be releasing Stable Diffusion support under Hugging Face's Diffusers library \cite{hf}. For more examples of Lego with various subjects, please see Figure \ref{fig:additional-results} in Supplementary material.

We tested Lego's concept inversion against natural language for 10 concepts, generating 200 images per concept using Lego and LDM with language control. Three concepts specified cardinality (3, 4, or 5 subjects). Participants counted subjects in each image. Lego consistently outperformed natural language guidance in producing correct subject counts (see Table \ref{tab:quantllm}). For the remaining seven concepts, we performed two different studies. A Visual Question Answering (VQA) large language model (Flan-T5 XL) \cite{ramesh2022hierarchical} was used to answer questions about 200 images generated using Lego, versus 200 images generated using natural language descriptions of each concept (2800 total images). Our experiments showed that VQA models performed better when asked ``Yes'' or ``No'' questions about specific concepts rather than asking more general questions such as `` What is this image?''. For instance, we generated images of ``toy bears frozen in ice''. We then prompted the VQA model with the question: `` Is the toy bear frozen in ice?''. Table \ref{tab:quantllm} shows the number of images where the VQA model confirms the concept is in the image. Consistent with the numerical concepts, 
Lego performs better than natural language for all concepts. In Supplementary Section \ref{supp:design}, we give a detailed overview of the prompts used and the generated images.

\par While VQA suggests Lego outperforms LDM, a more comprehensive evaluation for generative models, especially concerning general concepts, involves human preference metrics. To this end, we used Amazon Mechanical Turk \cite{mturk} and for each concept, paired the 200 images generated by Lego with the other 200 generated by LDM. We asked users to select the image they think best represents the concept for each pair of images. Each question was answered by 10 users (total of 14000 answers for all 7 concepts). The majority vote determined the outcome, showing that Lego was preferred over LDM in over 70\% of cases (Table \ref{tab:quantllm}). While we use Lego to invert concepts, in Section \ref{supp:obj-reconstruct} we show that the learned subject embedding is also able to perform similar to TI in inverting the subject of the reference images.

\begin{table}[h]
\caption{LLM metric reports the number (out of 200) of images, where Flan-T5 XL model confirms the image contains the concept. The Human metric reports the number of images with correct subject cardinality for the numerical concepts, and the percentage of time users preferred one method's images compared to the other.}
    \centering
    \small
    \begin{tabular}{c|c|c|c|c}
         \multirow{2}{*}{Concept} & \multicolumn{2}{c|}{LLM $\uparrow$} & \multicolumn{2}{c}{Human $\uparrow$} \\
         \cmidrule(lr){2-3} \cmidrule(lr){4-5}
         & Lego & LDM & Lego & LDM \\
         \midrule
         3 & - & - & 107 & 85 \\ 
         4 & - & - & 63 & 18 \\ 
         5 & - & - & 36 & 18 \\ 
         \midrule
         Frozen in ice & 92 & 55 & 68.5\% & 31.5\%\\ 
         Burnt and melted & 136 & 61& 77\% & 23\% \\
         Closed eyes & 111 & 75 & 74.5\% & 25.5\%\\ 
         Smiley Emoji Face & 199 & 151 & 67.5\% & 32.5\% \\ 
         Crumpled and squeezed & 147 & 64& 60.5\% &39.5\% \\ 
         Walking on a rope & 133 & 4 &  84 \%& 16\%\\ 
         Arms raised & 124 & 49 & 70\% & 30\%\\
    \end{tabular}
    \label{tab:quantllm}
\end{table}

\subsection{Concept Composition and Complexity}
Lego's learned concept embeddings can be combined, as demonstrated in Figure \ref{fig:composition} (left). It seamlessly composes concepts like ``closed eyes'' from cat images and ``walking on a rope'' from figure images. This enables creating scenes like a lion walking on a rope with closed eyes. While we have showcased Lego's ability with mostly two-word-embedding concepts, the synthetic example in Figure \ref{fig:composition} (right) illustrates its capability to invert more complex scenarios, learning and applying ``walking on a rope with closed eyes'' from a single example to the subject ``lion''.

\subsection{Ablation Study}
\label{sec:ablation}
\begin{figure}
\centering
    \includegraphics[width=0.9\linewidth]{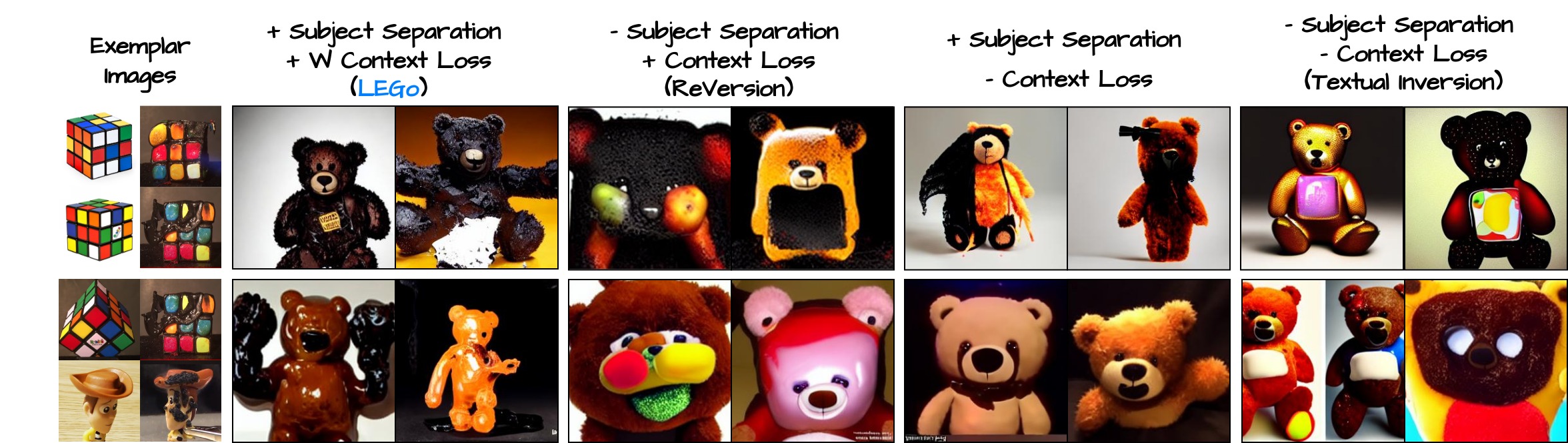}
\caption{\label{fig:ablation} This figure shows examples of our ablation study for learning the concept of \textit{``burnt and melted''} from single subject (top row) and multi subject (bottom row) example images and transferring it to a \textit{``Toy bear''}. We synthesized 100 images for each ablation category. You can find all 100 images for each category in the Supplementary Section \ref{supp:ablation}. Please zoom in to see the details in the images.}
\end{figure}

We studied the effect of \textit{\textbf{Subject Separation}} and \textit{\textbf{Context Loss}} on inverting three concepts; \textit{``burnt and melted''}, \textit{``closed eyes''} and \textit{``walking on rope''}. TI requires a few images of a subject or subjects in the same style (e.g., a few Monet paintings), while ReVersion needs multi-subject images for relation inversion. For the \textit{``burnt and melted''} concept, we used both single and multi-subject reference images to cover various settings. Our ablation study compares Lego, incorporating both Subject Separation and Context Loss, with three other combinations that remove one or both of these steps. Note that single / multi-subject experiments with neither Subject Separation nor Context Loss resemble performing a regular Textual Inversion \cite{t2i}. Performing single / multi-subject experiments by adding the Context Loss without Subject Separation resembles the ReVersion framework \cite{huang2023reversion}. Figure \ref{fig:ablation} shows the ablation results for \textit{``burnt and melted,''} highlighting the impact of Subject Separation and Context Loss for concept inversion. Complete ablation results are shown in Supplementary Section \ref{supp:ablation}.

\subsection{Example Image Size + Choice of Positive / Negative Words}
In Section \ref{supp:size} of the supplementary, we show the effect of increasing reference images on Lego's performance. Our approach to selecting positive and negative words 
involved choosing synonyms and antonyms of semantically 
meaningful words for the concept (Figures \ref{supp:num-template} and \ref{supp:7-template}), demonstrating 
robustness without requiring adjustments. While studying 
word selection’s subjective impact on inversion is challenging, it remains an interesting avenue to be explored

\section{Conclusion}
\label{sec:conclusion}

In this work, we took a first look at capabilities of T2I models in synthesizing adjective and verb concepts by  reference examples. We showed that entanglement of such concepts with a subject and the need for using multiple word embeddings in describing more complex concepts hinders current inversion methods. We proposed Lego that effectively inverts such personalized concepts from as few as 4 example images by disentangling concepts from subjects with a \textit{Subject Separation} step and further enables defining concepts with multiple embeddings by using a \textit{Context Loss} that guides each embedding towards a meaningful place in the textual embedding space.

\vspace{-2mm}\subsubsection{Acknowledgements.} This research was partially funded by the Ministry of Education and Science of Bulgaria (support for INSAIT, part of the Bulgarian National Roadmap for Research Infrastructure).
\newpage
{
    \small
    \bibliographystyle{splncs04}
    \bibliography{main}
}
\clearpage
\newpage
\setcounter{page}{1}
\begin{center}
\section*{Lego: Learning to Disentangle and Invert Personalized Concepts Beyond Object Appearance in Text-to-Image Diffusion Models - supplemental document}
\end{center}

\begin{itemize}
    \item We go over Lego's limitation and ethical use of personalized generative models in Section \ref{limits}
    \item Please refer to Section \ref{supp:size} to see the effect of increasing the number of reference images for performing Lego.
    \item In Section \ref{supp:obj-reconstruct} we show that the learned subject embedding by Lego can also be used and it performs similar to TI.
    \item To see \textbf{Lego inverted concepts applied to more subjects}, please see Figure \ref{fig:additional-results}.
    \item If you are interested in \textbf{details of the experimental design} such as text templates and how $\mathcal{P}$ and $\mathcal{N}$ were chosen for each concept, please refer to Section \ref{supp:design}.
    \item For details on \textbf{the prompts used in generating concepts using the baseline and details on the VQA study}, please see Section \ref{supp:results}.
    \item 200 images generated using Lego and 200 images generated using the baseline are shown in Figures \ref{supp:lego-200-frozen} to \ref{supp:text-200-walk}.
    \item We show how ELITE \cite{wei2023elite} and ControlNet \cite{zhang2023adding} perform in inverting general concepts in Section \ref{supp:cnet-elite}.
    \item The full results for our ablation study can be found in Section \ref{supp:ablation}.
    
\end{itemize}

\section{Limitations and Ethical Statement}
\label{limits}
Lego struggles to invert concepts beyond the capabilities of the backbone, like facial expressions with earlier versions of Stable Diffusion. Our research is dedicated to ethical and responsible data use, emphasizing the need for socially responsible applications in personalized visual media.

\section{The Effect of Dataset Size}
\label{supp:size}
In our experiments, we only used 4 images per experiment; 2 reference images of the subject only and 2 reference images of the concept applied to the subject. 
We trained Lego for the "closed eyes" and "raised arms" concept with 2, 4 \& 8 concept images and 2 subject-only images. We then generated 100 images for each concept. We then used our LLM-VQA evaluation setting to evaluate Lego's performance in each of these settings. For the concept of "closed eyes", Lego achieved an accuracy of 75\% when 2 concept images were shown to Lego. This increased to 83\% for both experiments with 4 and 8 concept reference images. Similarly, for the concept of "raised arms", an accuracy of 78\% was achieved with only 2 concept images. Using 4 and 8 concept images increased the accuracy to 80\% and 81\% respectively.

\section{Lego's Ability in Subject Synthesis}
\label{supp:obj-reconstruct}
\begin{figure}[h]
\centering
\includegraphics[width=\linewidth]{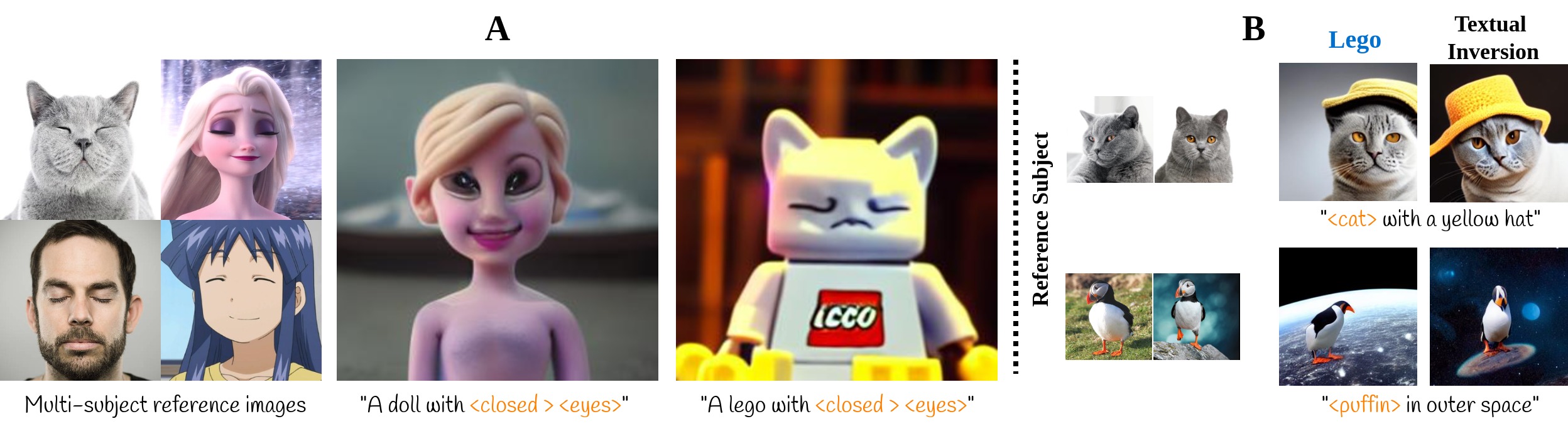}
\caption{(\textbf{A}) Appearance leakage in TI for inversion of ``closed eyes" while using multiple subject reference images. (\textbf{B}) Lego and TI show similar capabilities in synthesizing objects cat and bird.}
\label{fig:r2}
\end{figure}

While Lego's main objective is to learn a good representation for synthesizing a concepy, it does so with a Subject separation step that also learns an embedding to represent the sibject. We conducted experiments to see whether the learned subject embedding performs as well as TI in synthesizing the subject. We generated 40 images of 2 subjects with TI and Lego. In a user preference study with 5 participants, TI was preferred 55\% of the time over Lego (45\%), showing similar capabilities in generating the subject with TI. Two such examples are shown in Figure \ref{fig:r2}.

\vspace{0.9em}\section{Additional Results}
\label{supp:add-results}
\begin{figure}[ht]
\centering
    \includegraphics[width=\textwidth,height=0.67\textheight,keepaspectratio]
{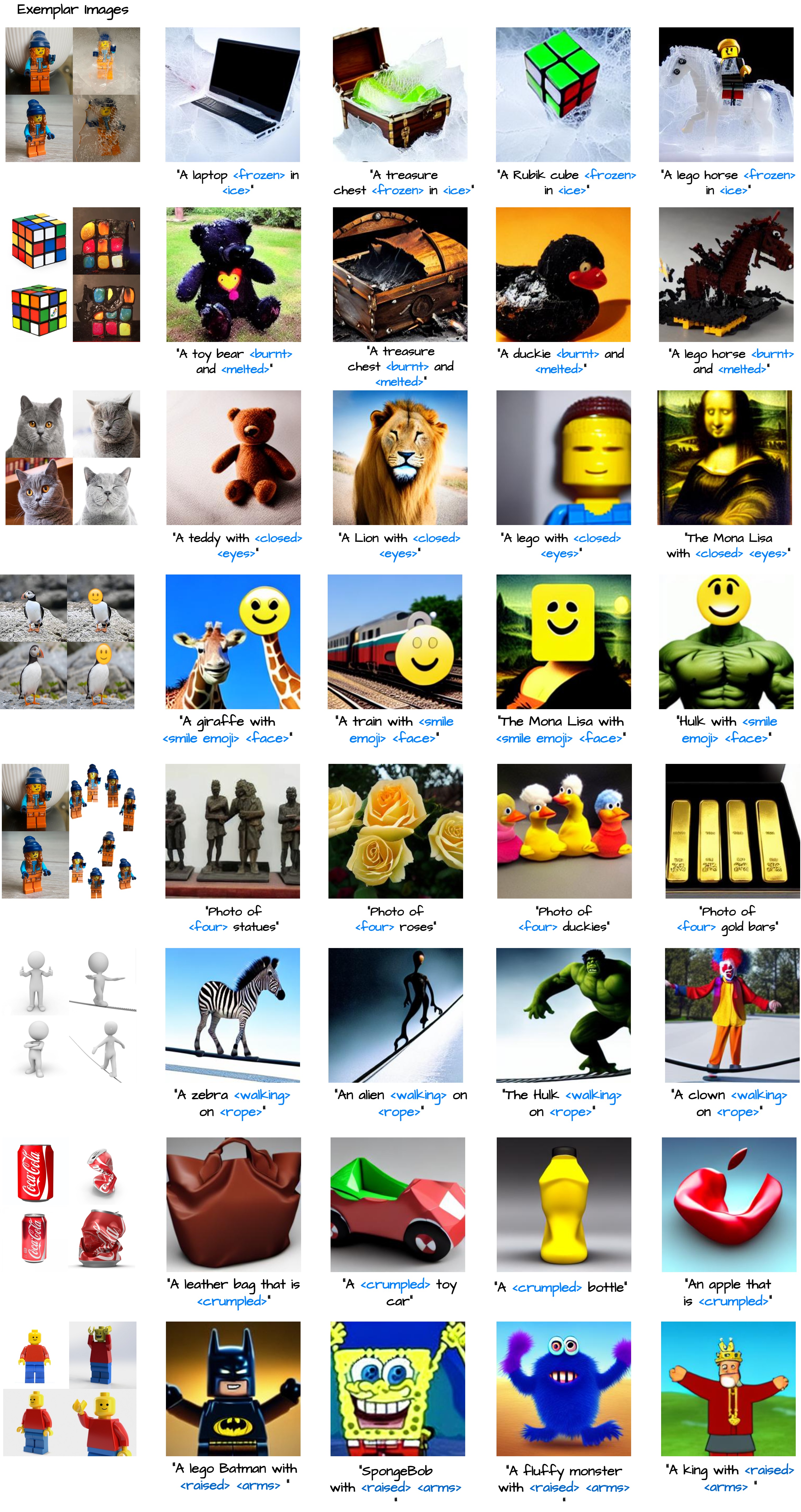}
\caption{\label{fig:additional-results} We show Lego learned concepts applied to more subjects. please zoom in to see details of images.}
\label{supp:more-results}
\end{figure}

In Figure \ref{supp:more-results}, we show the results of Lego learned concepts, applied to four additional subjects.

\section{Experiment Design}
\label{supp:design}
In this section, we go over the design of each concept inversion experiment, including the template of prompts used and selecting the appropriate positive ($\mathcal{P}_{i}$) and negative ($\mathcal{N}_{i}$) set of words for each concept embedding.

\subsection{Numerical Concepts}
\begin{figure*}[h]
\centering
    \includegraphics[width=\linewidth]{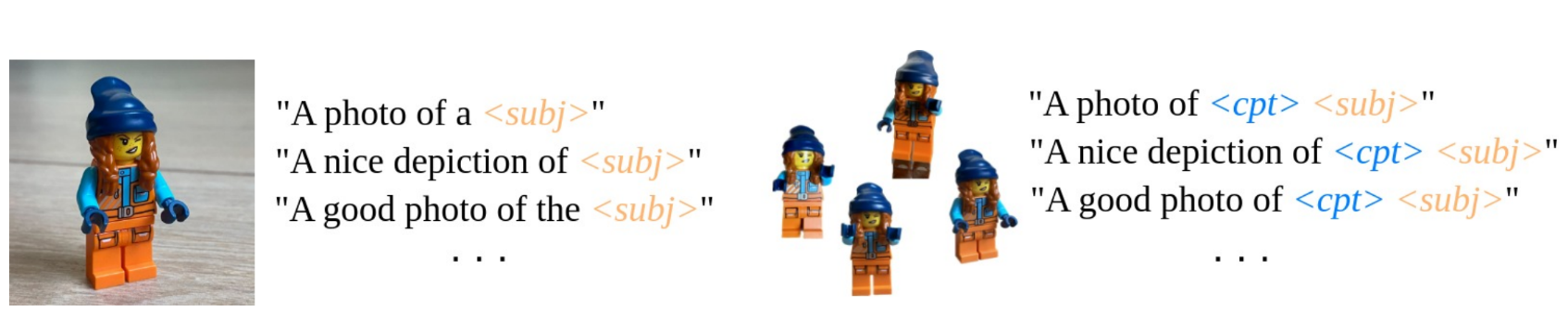}
\caption{Sample templates for inverting numerical concepts. Single subject images will be make up $\mathcal{I}_{\overline{C}}$ and the templates will only have the $<$\textit{subj}$>$ (left) while images with as many subjects as we are trying to invert the number for will make up $\mathcal{I}_{C}$ with the template having both $<$\textit{subj}$>$ and $<$\textit{cpt}$>$ embeddings (right).}
\label{supp:num-template}
\end{figure*}

In our work, we inverted concept numbers 3, 4 and 5 by using example images of 1 Lego figurine as $\mathcal{I}_{\overline{C}}$ and images of 3, 4, and 5 Lego figurines as $\mathcal{I}_{C}$, respectively. Figure \ref{supp:num-template} shows sample text templates used for inverting such concepts. We dedicated an embedding $<$\textit{subj}$>$ for learning the subject appearance and an embedding $<$\textit{cpt}$>$ to learn the concept. For all numerical concepts, a single concept embedding was used.

To form the positive and negative sets of words that were used to calculate our context loss, we constructed the positive set $\mathcal{P}$ to be the digit and word description of the desired number and further constructed the negative set $\mathcal{N}$ to be the digit and word descriptions of a few neighbouring numbers. For instance, to learn the concept of $4$ from the Lego figurines in Figure \ref{supp:num-template}, we set $\mathcal{P}=\{\text{four}, \text{4}\}$ and $\mathcal{N}=\{\text{2, 3, 5, 6, two, three, five, six}\}$. 

\subsection{Remaining 7 Concepts}
\begin{figure*}[h]
\centering
    \includegraphics[width=0.88\linewidth]{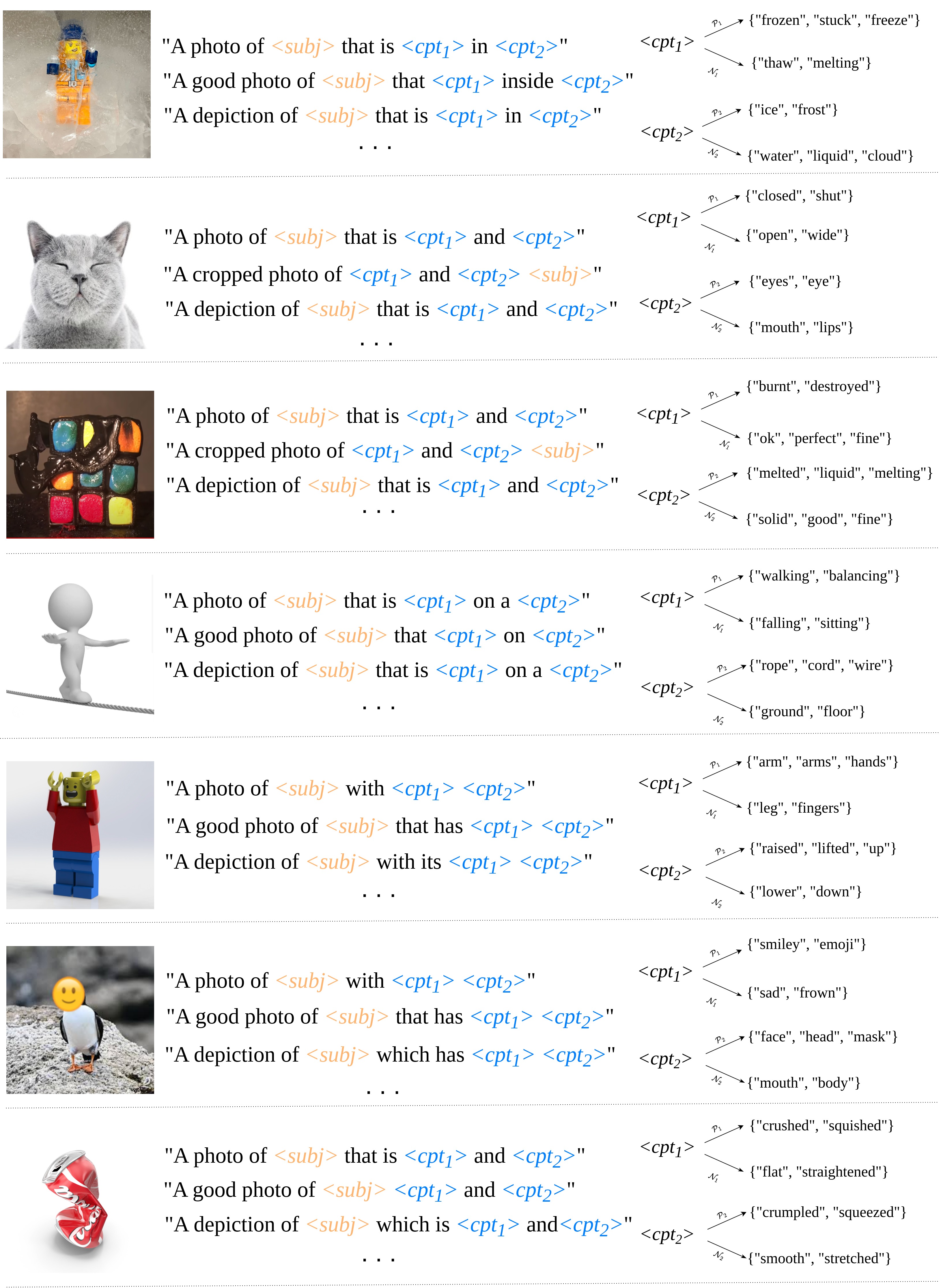}
\caption{Template samples for the non-numerical concepts explored in this work. All 7 concepts use two word embedding and we show the positive and negative words for each embedding.}
\label{supp:7-template}
\end{figure*}

In Figure \ref{supp:7-template}, we show template samples for the remaining 7 concepts that were explored in this paper. We skip the subject-only templates as they follow the same structure as the one we showed in Figure \ref{supp:num-template}. To the right of each concept's template, we provide each embedding's corresponding $\mathcal{P}$ and $\mathcal{N}$ that we used in our experiments.
\par In order to select $\mathcal{P}$ and $\mathcal{N}$, we used a few synonyms for the effect we are trying to invert. In the case of numerical concepts, our negative set was the neighbouring numbers due to the fact that they are in close embedding distance to the concept number. This is however not the case in many examples. For instance, for the concept `` closed eyes'', the positive set for one of the embeddings has the word ``closed''. The antonym word ``open'' however is not close to the embedding of ``closed''. Our experiments showed that in such cases, the negative set's words is not as important as the positive set, yet one should still select words based on the antonym scheme as selecting arbitrary words can cause the embedding to diverge from reaching a suitable place in the text embedding space.
\par We suggest following our sanity check step when using Lego to invert arbitrary concepts; during optimization, for each concept embedding, we monitor the 10 most and least similar words to the embedding. This allows you to see if your embedding is indeed getting close to a set of words that describe the concept. They could also suggest words for building the negative set if the embedding is also getting close to undesired words that you had not thought of before. 

\section{LLM and Human Preference Study}
\label{supp:results}
In this section, we provide the 200 images generated with Lego and LDM for each concept in our study. Both the VQA LLM (Flan) and Human preference (Mechanical Turk) study preferred Lego generated concepts over the baseline (LDM with natural language input). 

\par\noindent For our evaluation with the VQA model, we experimented with asking different types of questions based on the ground truth exemplar images. ``Yes'' or ``No'' questions about the concept being present in the image or not led to the highest correct answers based on the example images. Hence, we asked ``Yes'' or ``No'' questions about each image and according to the LLM, Lego generated concepts had more images aligned with each concept compared to the baseline, across all concepts (see Table \ref{tab:quantllm}).

\par
\begin{figure*}[th]
\centering
    \includegraphics[width=0.85\linewidth]{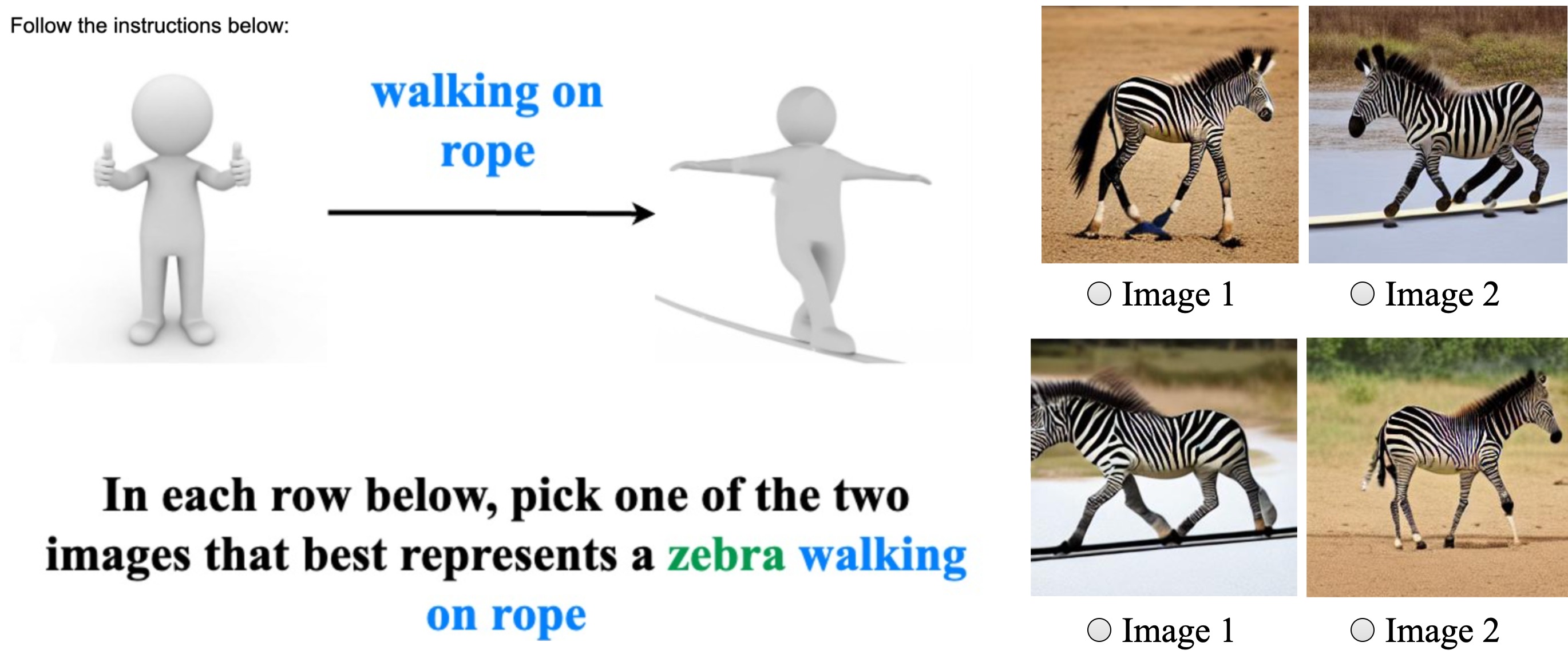}
\caption{The users were shown the image with instructions (left) on the concept they should be looking for in each pair of images. And below the instruction, 20 questions, each including 2 images (right). were given to users where they had to pick one of the two. The order of Lego and baseline generated concepts was randomized.}
\label{supp:mturk-temp}
\end{figure*}

\noindent For each concept in the Human preference study, 10 sets of questions were created, each containing 20 pairs of images, one generated using Lego's inversion of the concept synthesized through an LDM backbone and one generated using natural language description (see next paragraph for details) of the concept, synthesized through LDM. For each pair of images, users were prompted to pick one image between the two that best represents the concept (users were given both the exemplar images and text description of the concept for each question - see Figure \ref{supp:mturk-temp} for a snapshot of the user study interface for the concept of \textit{walking on rope}). We selected the image with majority user vote \cite{Rao2013WhatWO, Daniel2018QualityCI} for each pair as the preferred concept representation. Lego was preferred over the baseline in all concepts as shown in Table \ref{tab:quantllm}. 

\par\noindent In Figures \ref{supp:lego-200-frozen} to \ref{supp:text-200-walk} show the 200 images generated for each concept using Lego and natural language. With prompt tuning being an important aspect of getting the desired outcome from generative models, for each concept in the baseline experiments, we described the concept in 4 different natural language prompts, with LDM generating 50 images per prompt. Below we provide the prompts used for each concept.

\par\noindent We applied the concept of \textit{\textbf{frozen in ice}} to ``toy bear'' (see Figure \ref{supp:lego-200-frozen}) and we used the following prompts for LDM; 1) ``Photo of a toy bear frozen in ice'', 2) `` a toy bear stuck in a block of ice'', 3) ``a toy bear that is frozen in a block of ice'' and 4) ``photo of a toy bear that is stuck in ice'' (see Figure \ref{supp:text-200-frozen}).

\par\noindent We applied the concept of \textit{\textbf{burnt and melted}} to ``toy bear'' (see Figure \ref{supp:lego-200-melt}) and we used the following prompts for LDM; 1) ``Photo of a toy bear that is burnt and melted'', 2) ``photo of a melted and burnt toy bear'', 3) ``a toy bear that is burnt and charred'' and 4) ``photo of a burnt and destroyed toy bear'' (see Figure \ref{supp:text-200-melt}).

\par\noindent We applied the concept of \textit{\textbf{closed eyes}} to ``lion'' (see Figure \ref{supp:lego-200-eyes}) and we used the following prompts for LDM; 1) ``a lion with closed eyes'', 2) ``photo of a lion with its eyes closed'', 3) ``photo of a lion with its eyes shut'' and 4) ``photo of a lion that has closed eyes'' (see Figure \ref{supp:text-200-eyes}).

\par\noindent We applied the concept of \textit{\textbf{smiley emoji face}} to ``fluffy monster'' (see Figure \ref{supp:lego-200-emoji}) and we used the following prompts for LDM; 1) ``Photo of a fluffy monster with a smiley emoji face'', 2) ``photo of a fluffy monster with a yellow smiley emoji head'', 3) ``a fluffy monster with a yellow smiley emoji face'' and 4) ``photo of a fluffy monster having a yellow smiley emoji mask'' (see Figure \ref{supp:text-200-emoji}).

\par\noindent We applied the concept of \textit{\textbf{raised arms}} to ``toy bear'' (see Figure \ref{supp:lego-200-arms}) and we used the following prompts for LDM; 1) ``Photo of a toy bear with its arms raised'', 2) `` photo of a toy bear with arms up'', 3) ``photo of a toy bear with its hands up in the air'' and 4) `` photo of a toy bear having raised its arms'' (see Figure \ref{supp:text-200-arms}).

\par\noindent We applied the concept of \textit{\textbf{crumpled and crushed}} to ``toy car'' (see Figure \ref{supp:lego-200-crumple}) and we used the following prompts for LDM; 1) ``Photo of a toy car that is crushed and crumpled'', 2) `` photo of a crushed and destroyed toy car'', 3) ``photo of a crumpled toy car'' and 4) `` photo of a crushed and squeezed toy car'' (see Figure \ref{supp:text-200-crumple}).

\par\noindent We applied the concept of \textit{\textbf{walk on rope}} to ``zebra'' (see Figure \ref{supp:lego-200-walk}) and we used the following prompts for LDM; 1) ``Photo of a zebra walking on a rope'', 2) `` photo of a zebra balancing on a tightrope'', 3) ``a zebra walking on a tightrope'' and 4) `` photo of a zebra walking on a wire'' (see Figure \ref{supp:text-200-walk}).

\section{ControlNet and ELITE}
\label{supp:cnet-elite}
\begin{figure}[h]
\centering
    \includegraphics[width=0.5\linewidth]{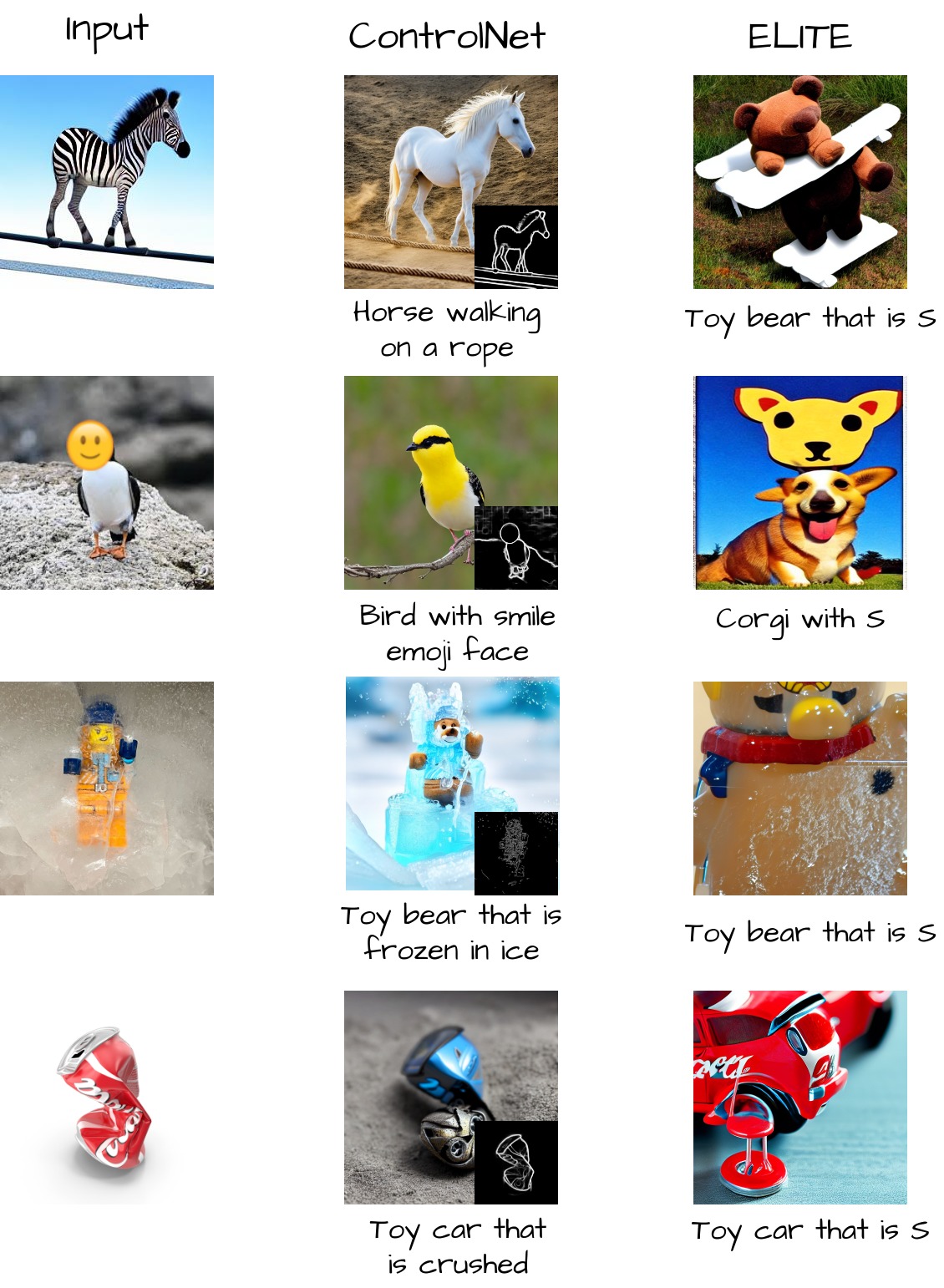}
\caption{\label{fig:cnet-elite} Results of ControlNet and ELITE for inverting concepts of \textit{walking on a rope, smiley emoji face, frozen in ice} and \textit{crushed and crumpled}}
\end{figure}

recall that ELITE \cite{wei2023elite} trained mapping networks on a large vision dataset in order to skip the embedding optimization step of Textual inversion by directly learning to map an image's subject to its corresponding embedding. Similar to TI, ELITE is an appearance based inversion method and also requires an input mask of the subject to be inverted. In Figure \ref{fig:cnet-elite} we show ELITE's attempt at inverting concepts of \textit{walking on a rope, smiley emoji face, frozen in ice} and \textit{crumpled and crushed}. 

\par\noindent ControlNet \cite{zhang2023adding} is a powerful method for adding conditional control to T2I Diffusion models; such as pose and layout. In Figure \ref{fig:cnet-elite} we condition Stable Diffusion on the edge map of example images of our concepts and prompt the model to follow these outlines. This is too restrictive for general concepts and as seen in the four concept examples, the results are not satisfactory. 

\section{All Lego and Baseline Images Used In User + LLM Study}
\label{supp:all}

\begin{figure*}[h]
\centering
    \includegraphics[width=0.61\linewidth]{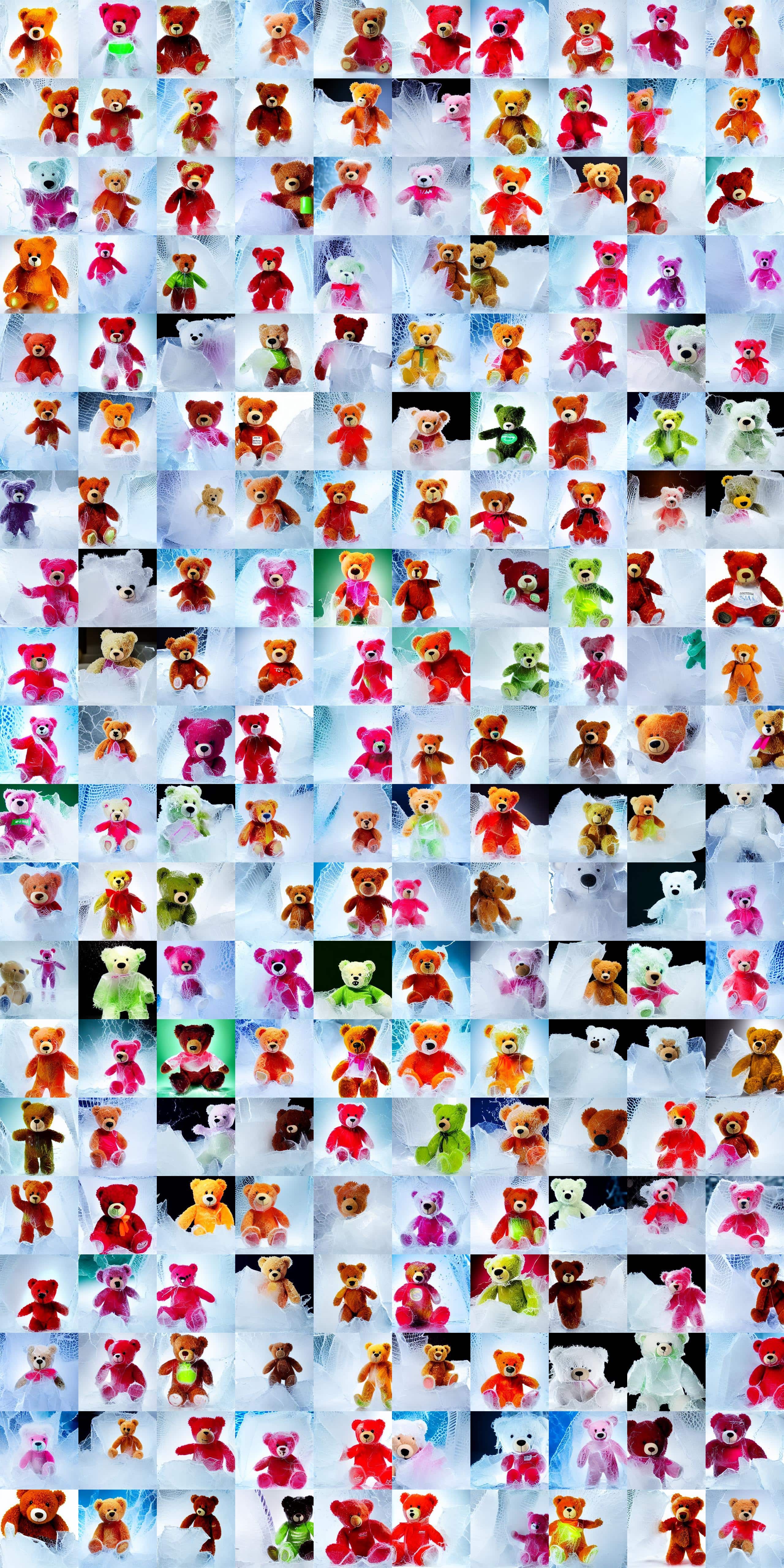}
\caption{\textbf{Lego} generated images with LDM backbone for the concept of \textit{frozen in ice}, learned from the Lego figurine frozen in ice example images, applied to a toy bear.}
\label{supp:lego-200-frozen}
\end{figure*}

\begin{figure*}[h]
\centering
    \includegraphics[width=0.61\linewidth]{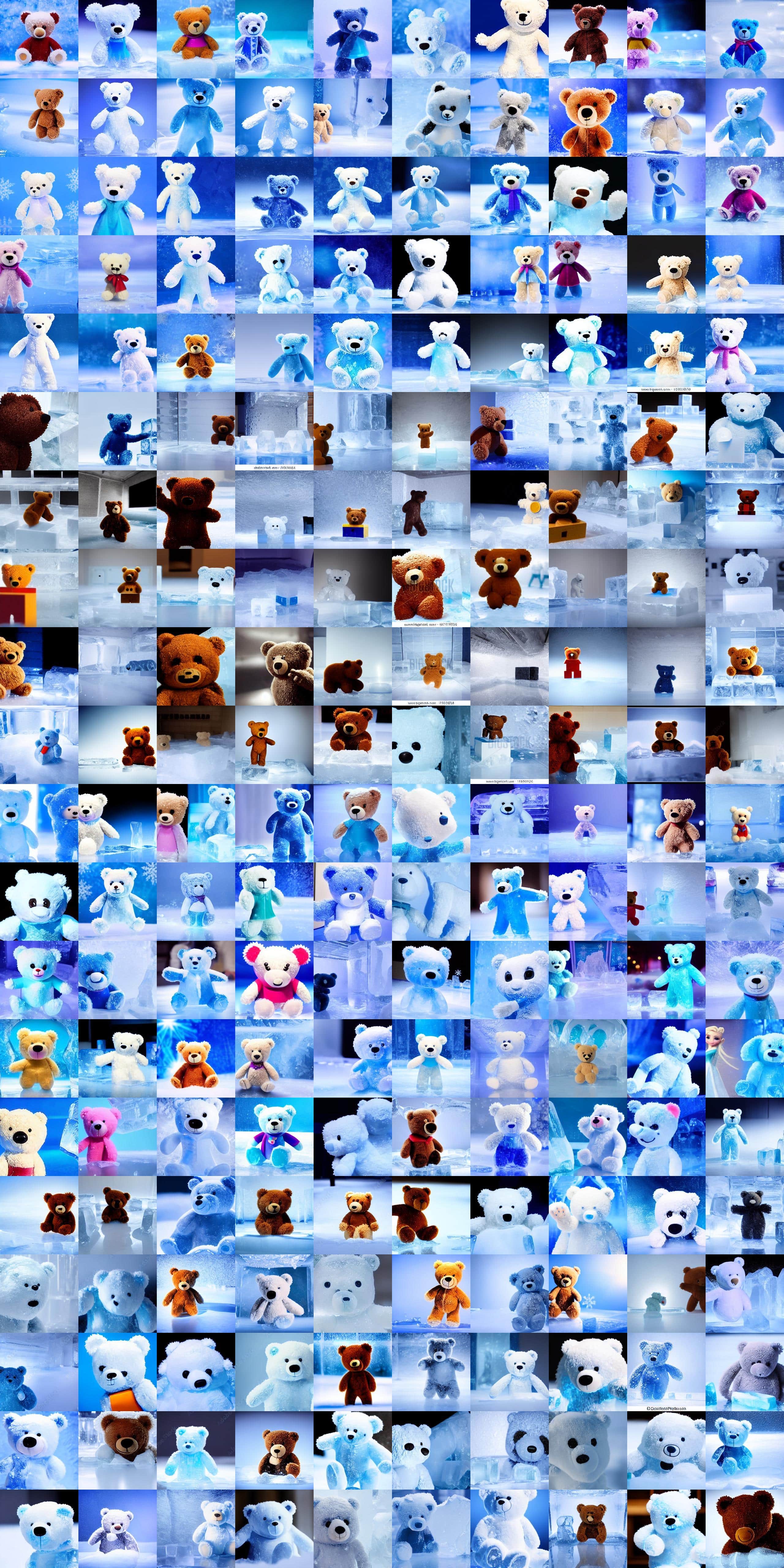}
\caption{\textbf{LDM} generated images using 4 different prompts, describing a toy bear that is frozen in ice.}
\label{supp:text-200-frozen}
\end{figure*}

\begin{figure*}[h]
\centering
    \includegraphics[width=0.61\linewidth]{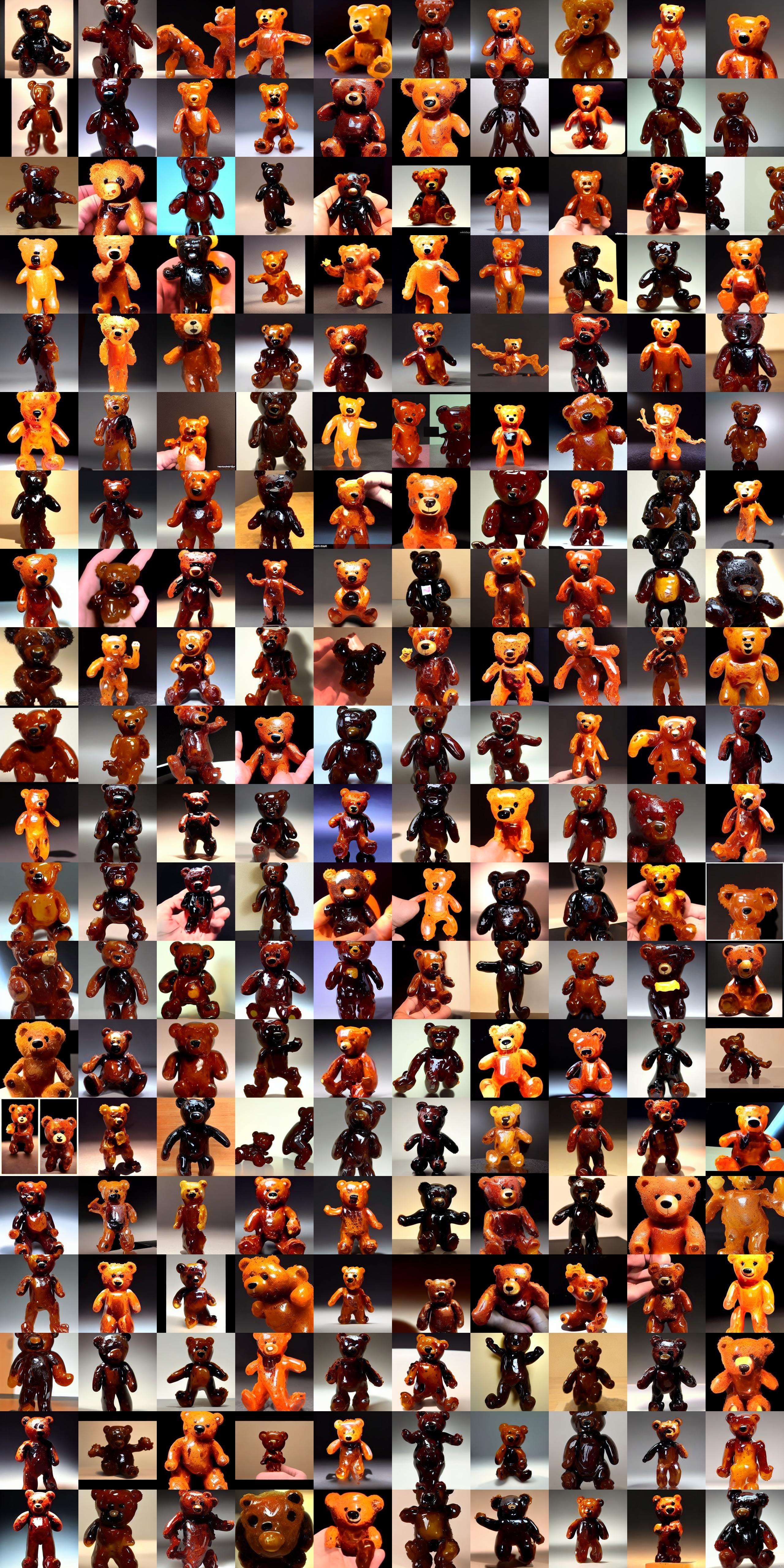}
\caption{\textbf{Lego} generated images with LDM backbone for the concept of \textit{burnt and melted}, learned from the burnt and melted Rubik's cube example images, applied to a toy bear.}
\label{supp:lego-200-melt}
\end{figure*}

\begin{figure*}[h]
\centering
    \includegraphics[width=0.61\linewidth]{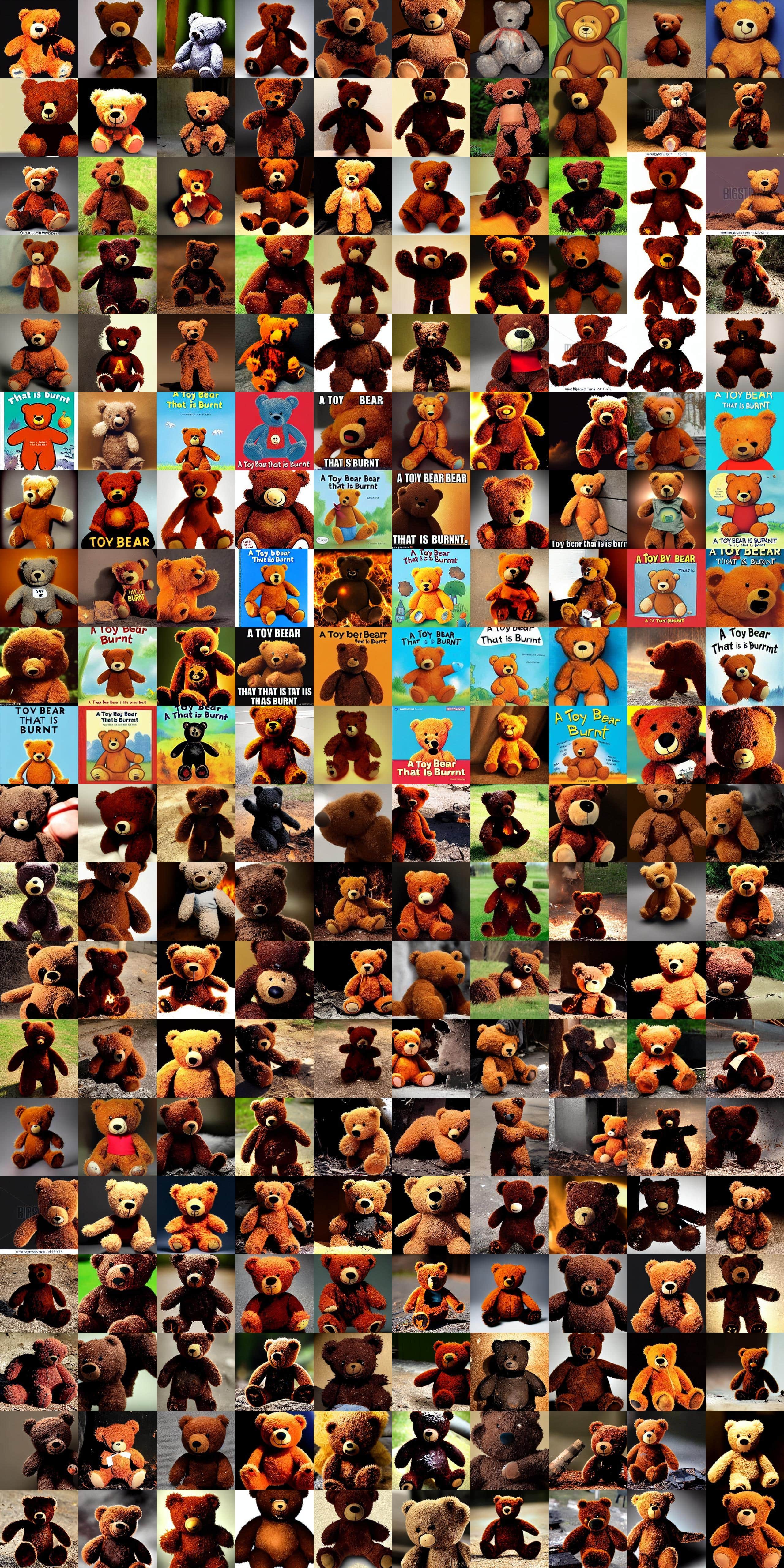}
\caption{\textbf{LDM} generated images using 4 different prompts, describing a toy bear that has been burnt and melted.}
\label{supp:text-200-melt}
\end{figure*}

\begin{figure*}[h]
\centering
    \includegraphics[width=0.61\linewidth]{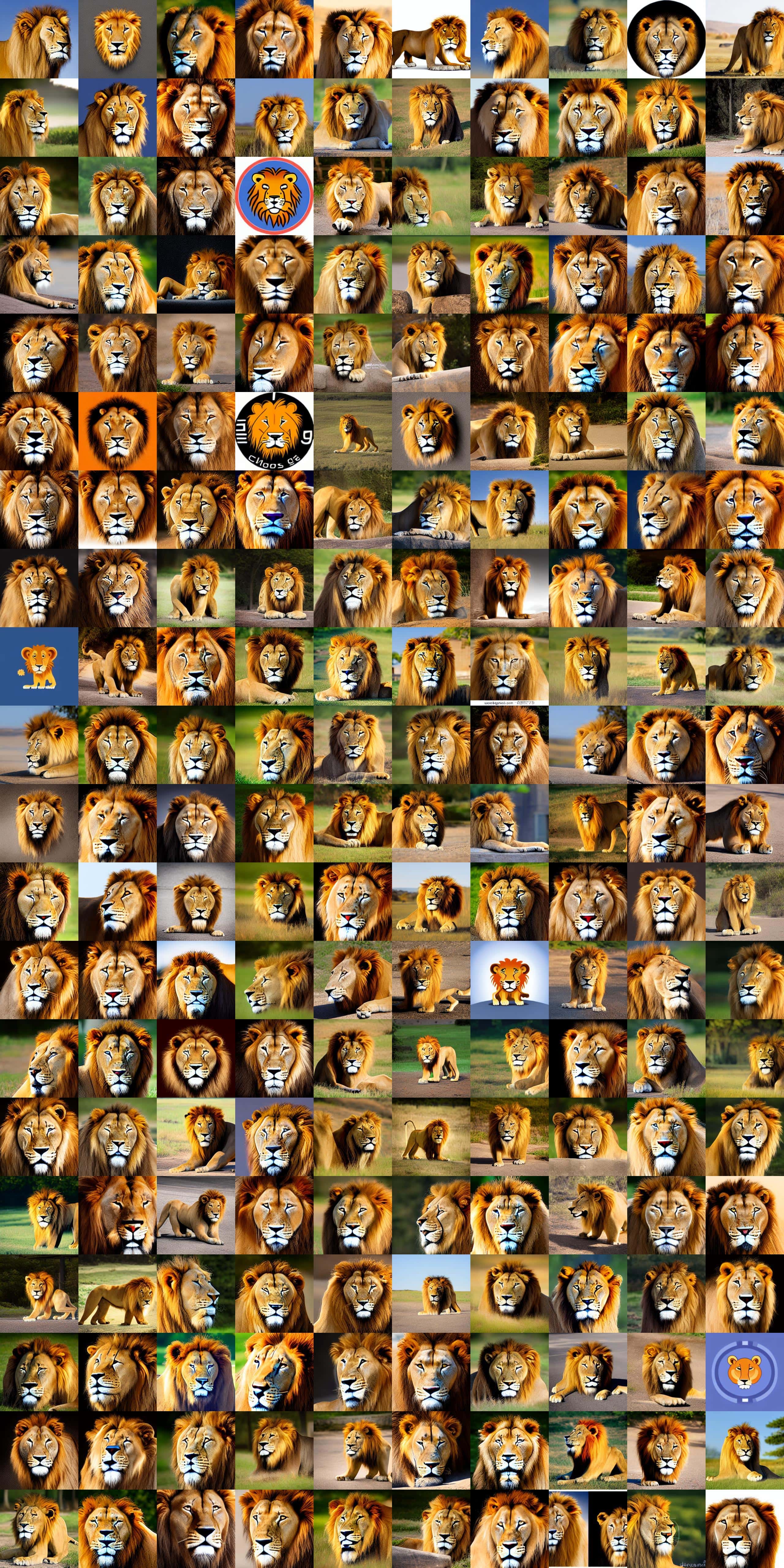}
\caption{\textbf{Lego} generated images with LDM backbone for the concept of \textit{closed eyes}, learned from the cat with closed eyes example images, applied to a lion.}
\label{supp:lego-200-eyes}
\end{figure*}

\begin{figure*}[h]
\centering
    \includegraphics[width=0.61\linewidth]{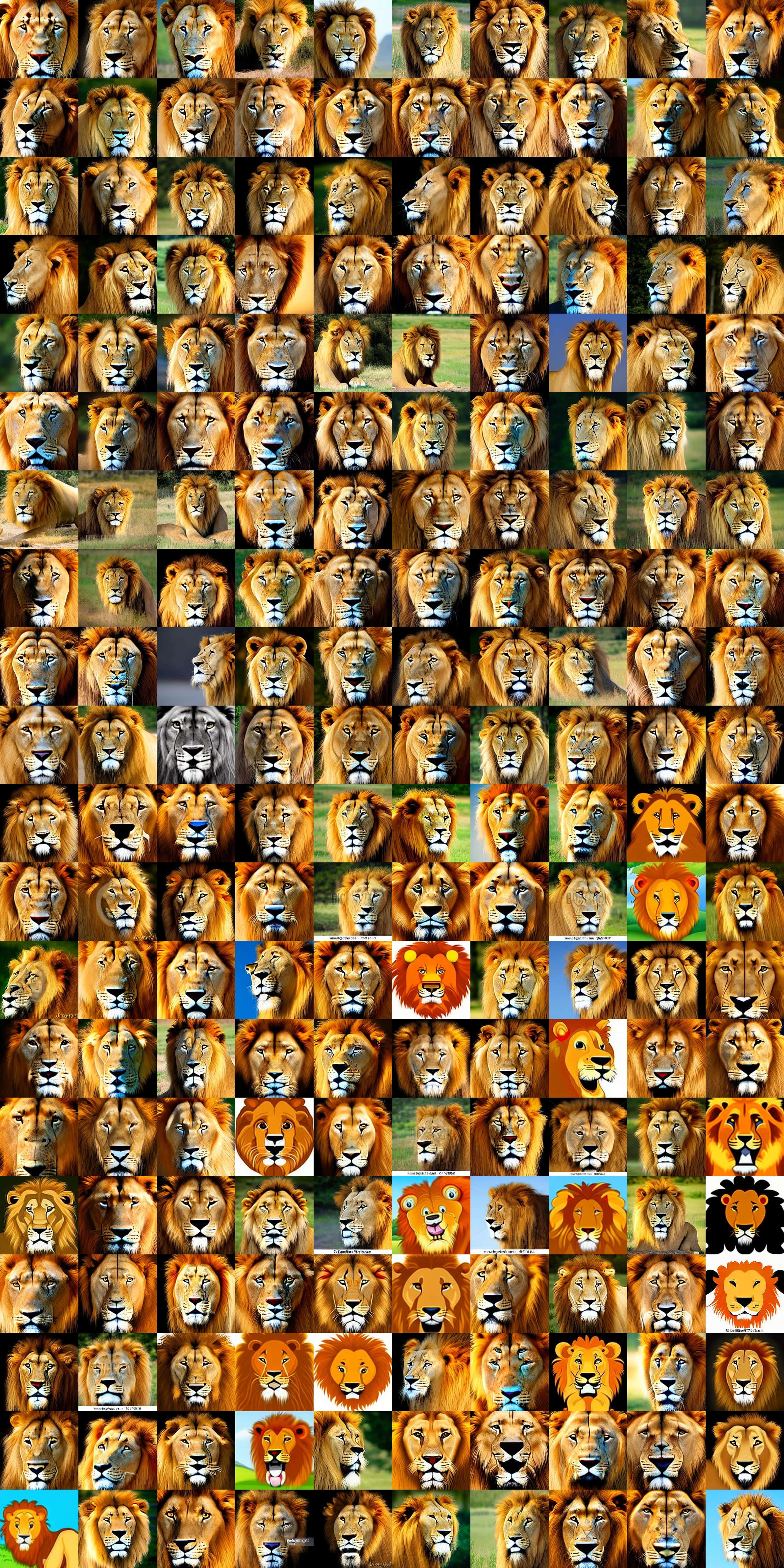}
\caption{\textbf{LDM} generated images using 4 different prompts, describing a lion with its eyes closed.}
\label{supp:text-200-eyes}
\end{figure*}

\begin{figure*}[h]
\centering
    \includegraphics[width=0.61\linewidth]{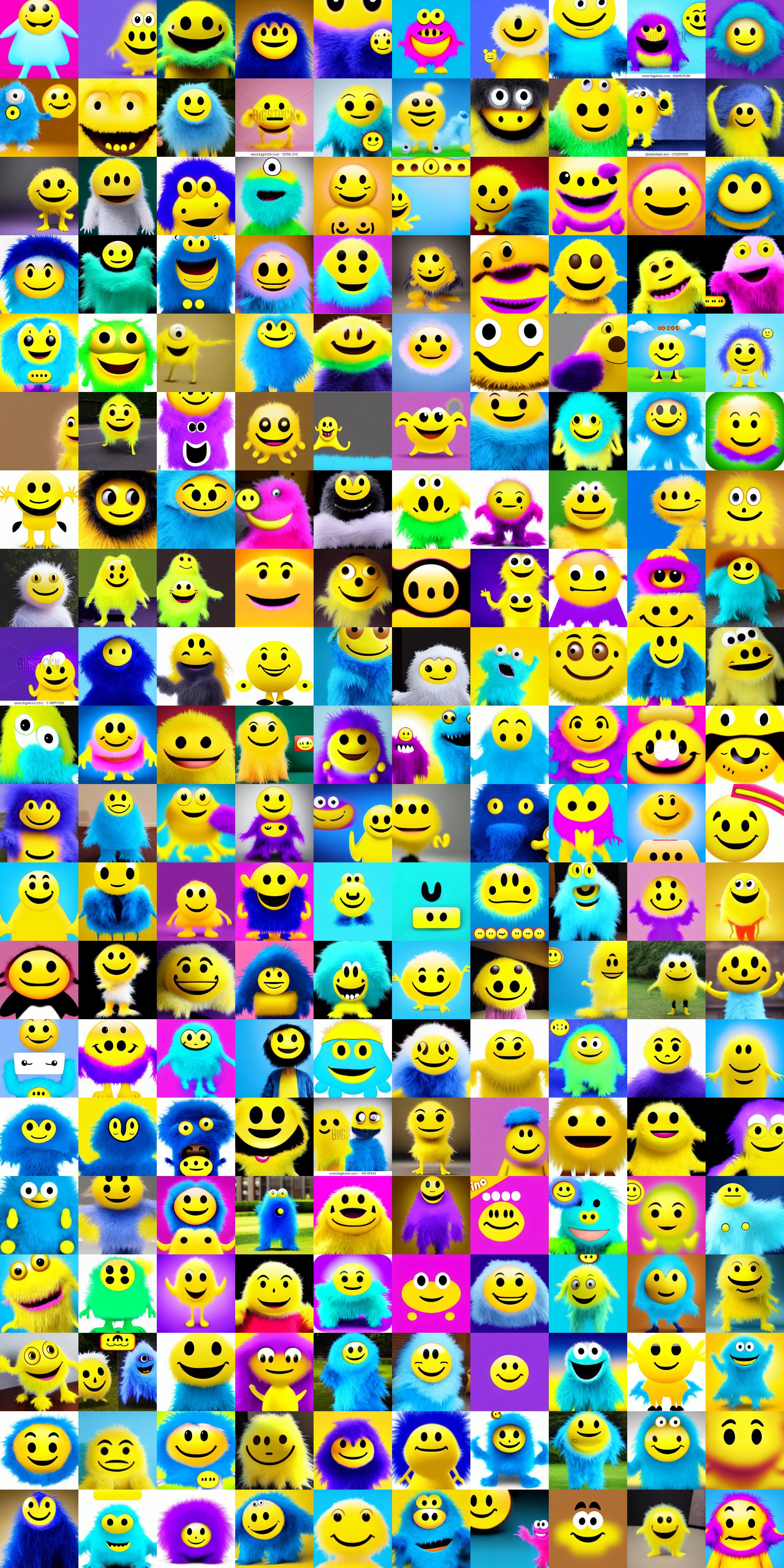}
\caption{\textbf{Lego} generated images with LDM backbone for the concept of \textit{smiley emoji face}, learned from the bird with smiley emoji face example images, applied to a fluffy monster.}
\label{supp:lego-200-emoji}
\end{figure*}

\begin{figure*}[h]
\centering
    \includegraphics[width=0.61\linewidth]{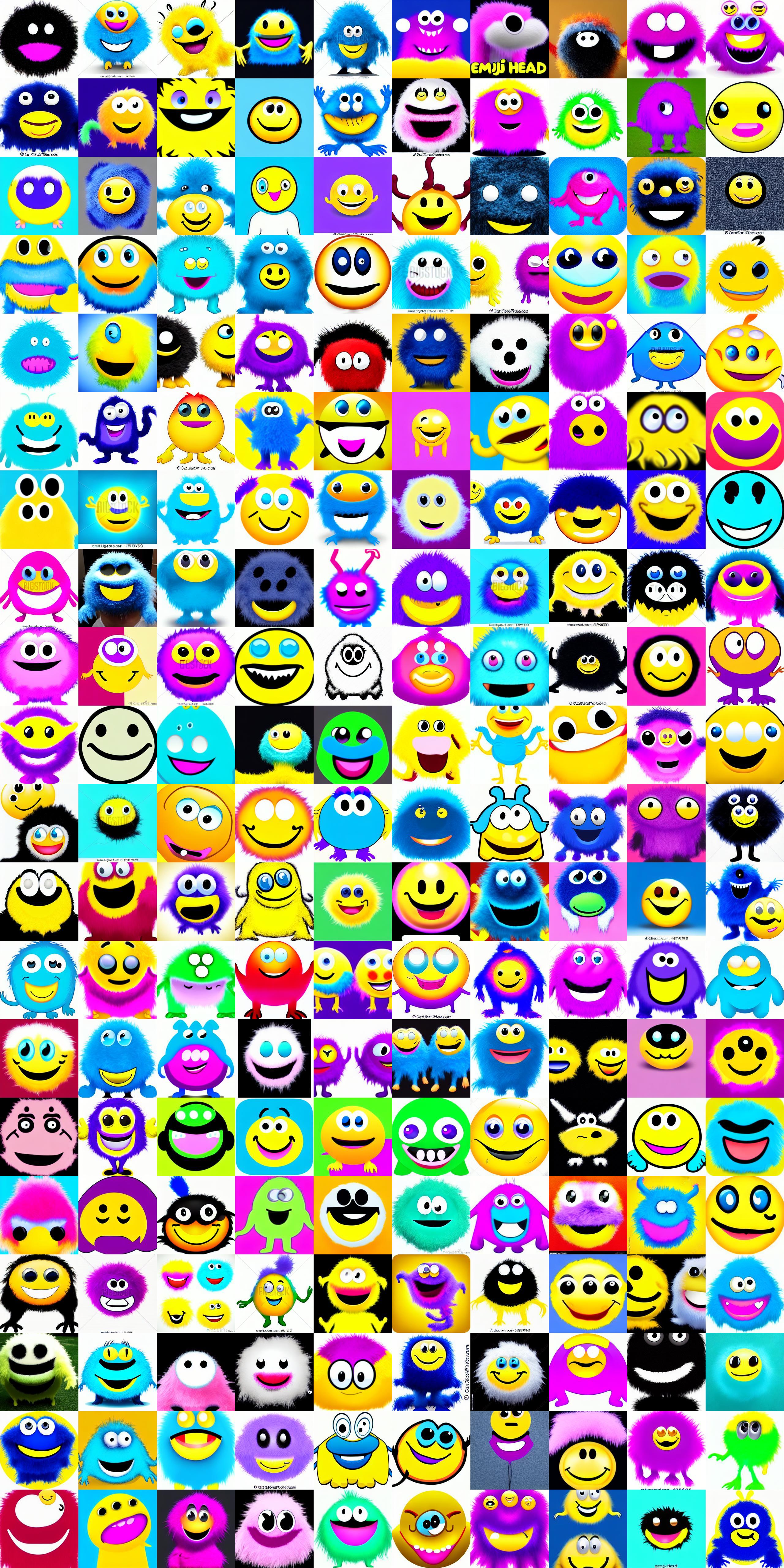}
\caption{\textbf{LDM} generated images using 4 different prompts, describing a fluffy monster with a yellow smiley emoji face.}
\label{supp:text-200-emoji}
\end{figure*}

\begin{figure*}[h]
\centering
    \includegraphics[width=0.61\linewidth]{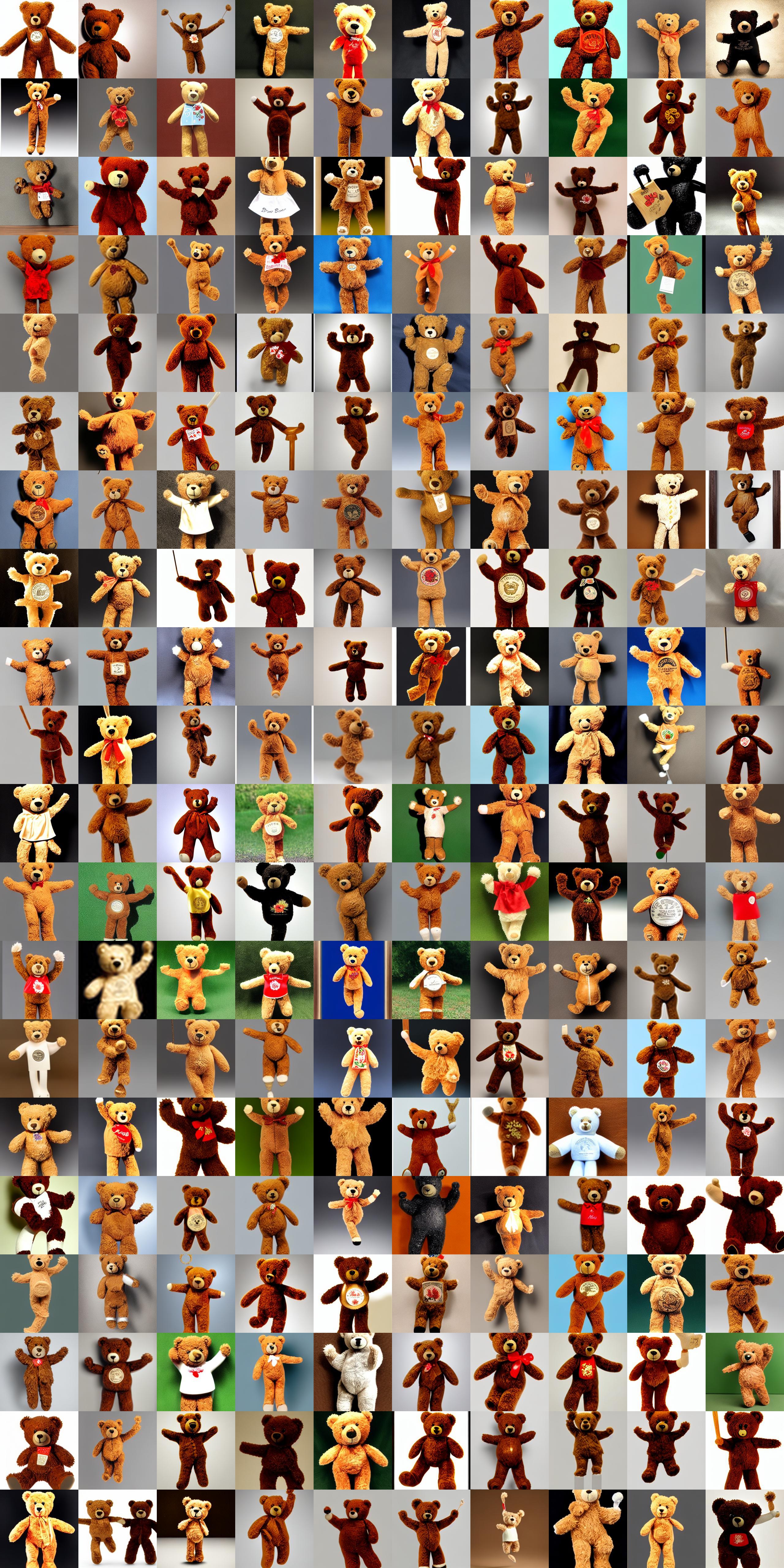}
\caption{\textbf{Lego} generated images with LDM backbone for the concept of \textit{arms raised}, learned from the Lego figurine with its arms raised example images, applied to a toy bear.}
\label{supp:lego-200-arms}
\end{figure*}

\begin{figure*}[h]
\centering
    \includegraphics[width=0.61\linewidth]{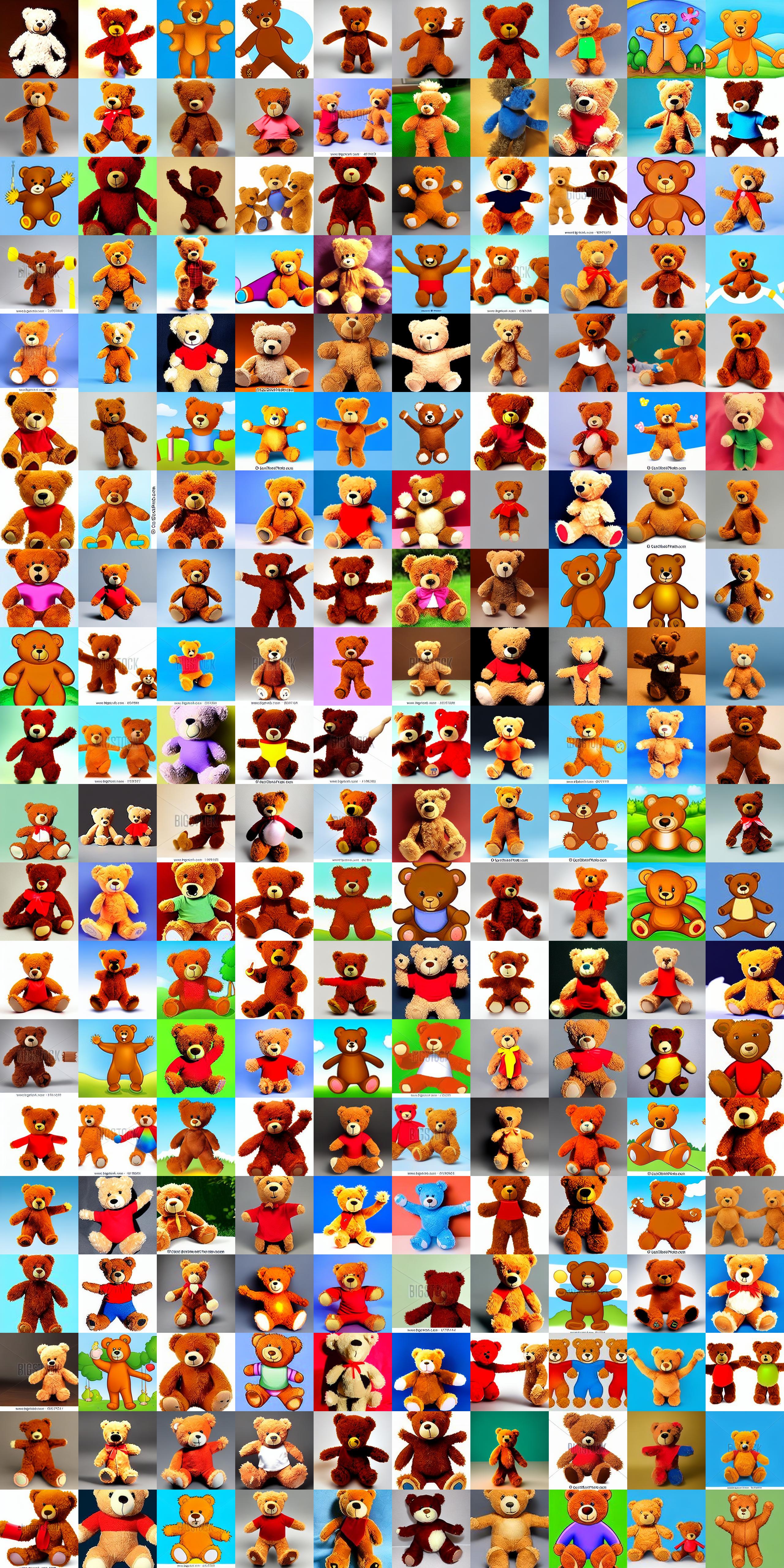}
\caption{\textbf{LDM} generated images using 4 different prompts, describing a toy bear with its arms raised.}
\label{supp:text-200-arms}
\end{figure*}

\begin{figure*}[h]
\centering
    \includegraphics[width=0.61\linewidth]{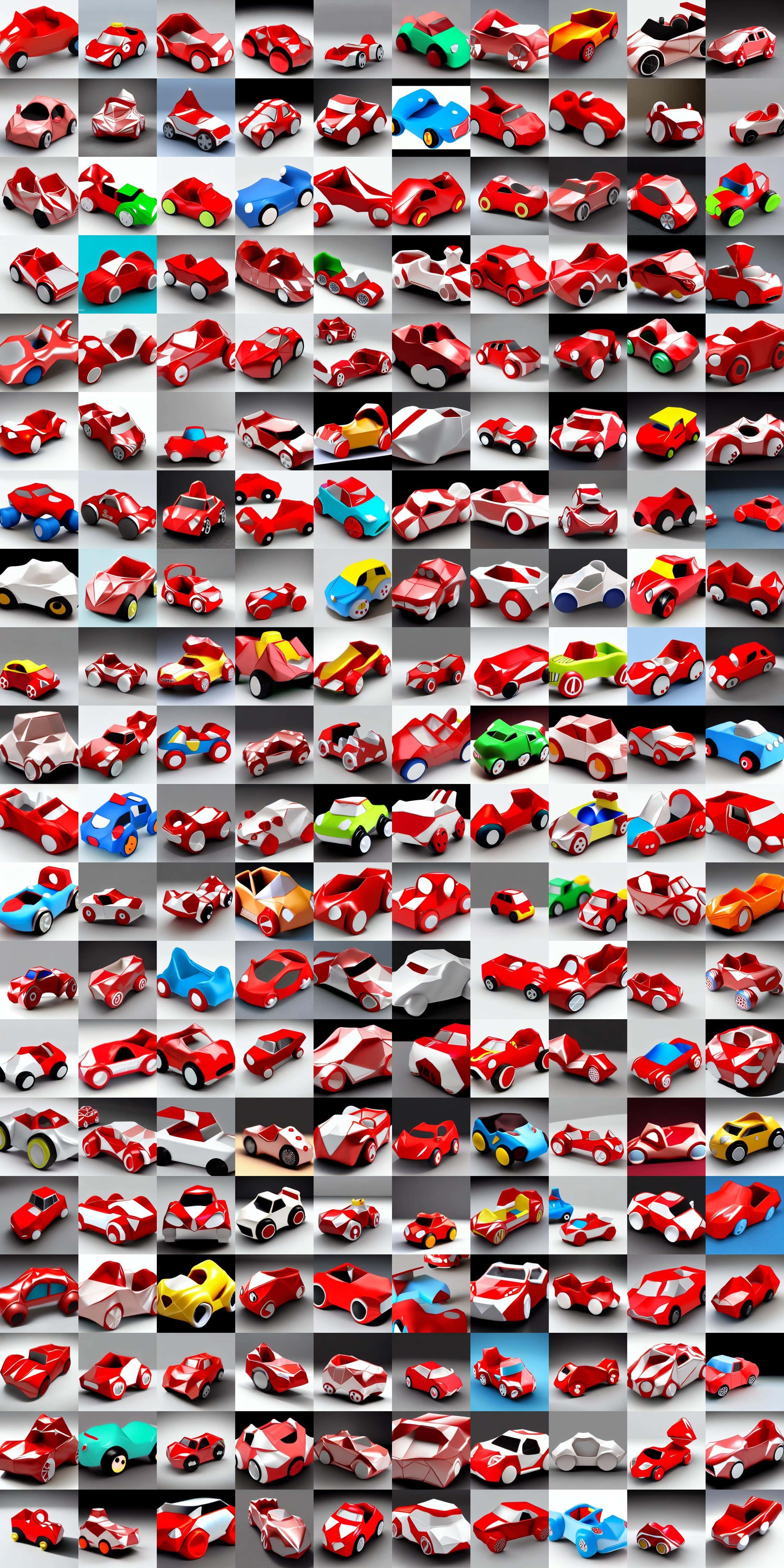}
\caption{\textbf{Lego} generated images with LDM backbone for the concept of \textit{crushed and crumpled}, learned from the crushed and crumpled soda can example images, applied to a toy car.}
\label{supp:lego-200-crumple}
\end{figure*}

\begin{figure*}[h]
\centering
    \includegraphics[width=0.61\linewidth]{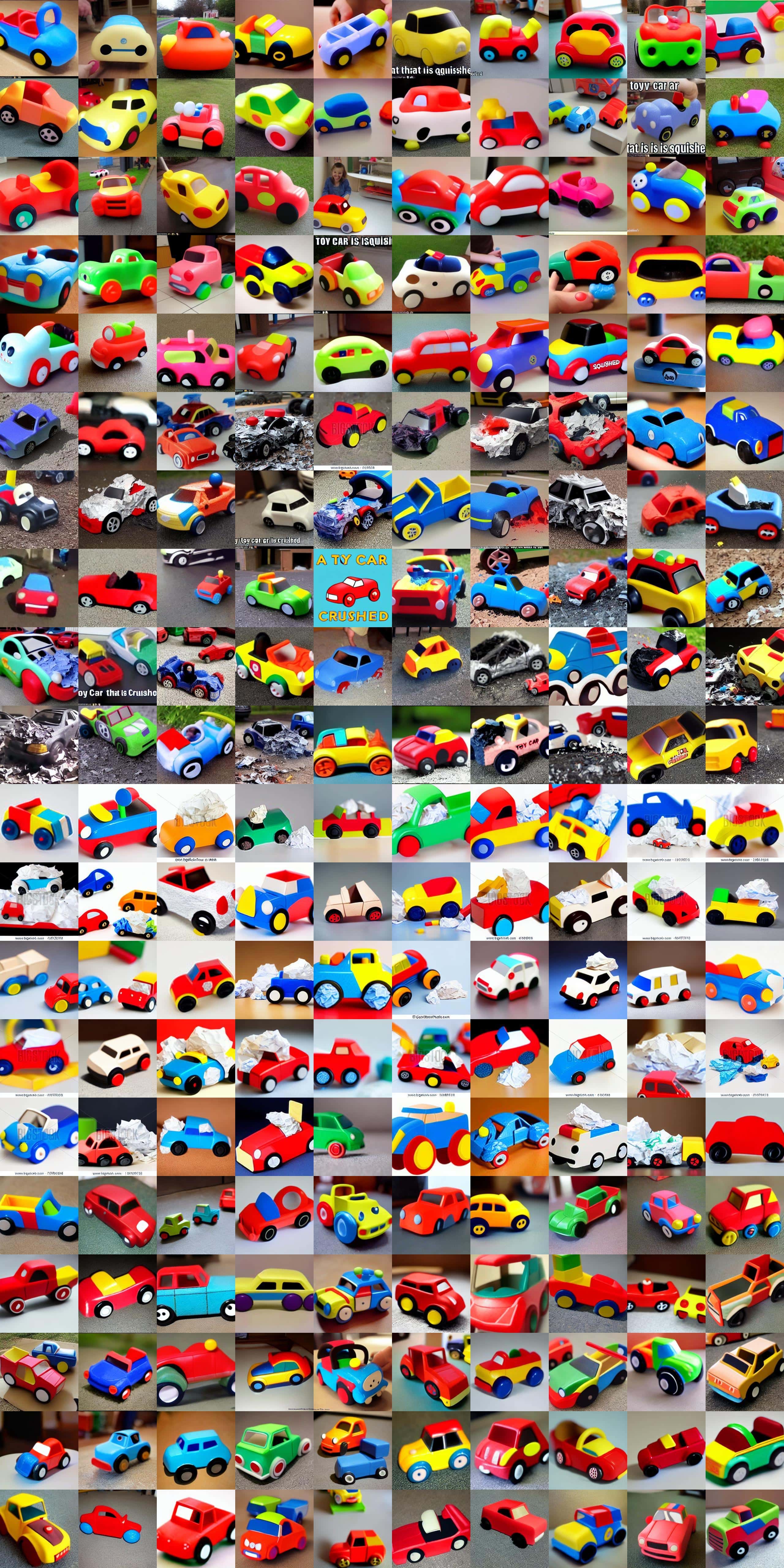}
\caption{\textbf{LDM} generated images using 4 different prompts, describing a toy car that is squished and crumpled.}
\label{supp:text-200-crumple}
\end{figure*}

\begin{figure*}[h]
\centering
    \includegraphics[width=.61\linewidth]{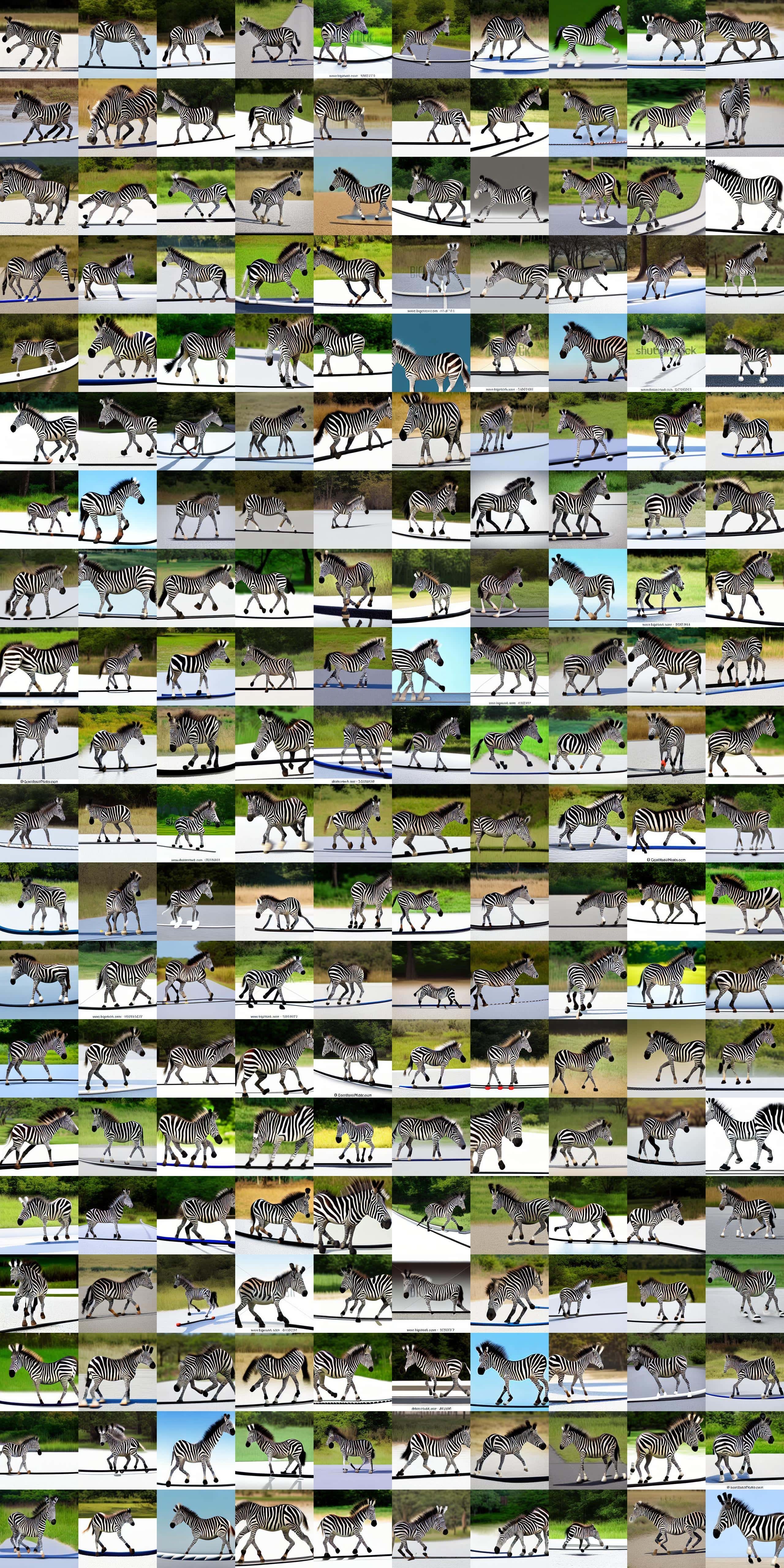}
\caption{\textbf{Lego} generated images with LDM backbone for the concept of \textit{walking on rope}, learned from the figure walking on a rope example images, applied to a zebra.}
\label{supp:lego-200-walk}
\end{figure*}

\begin{figure*}[h]
\centering
    \includegraphics[width=0.61\linewidth]{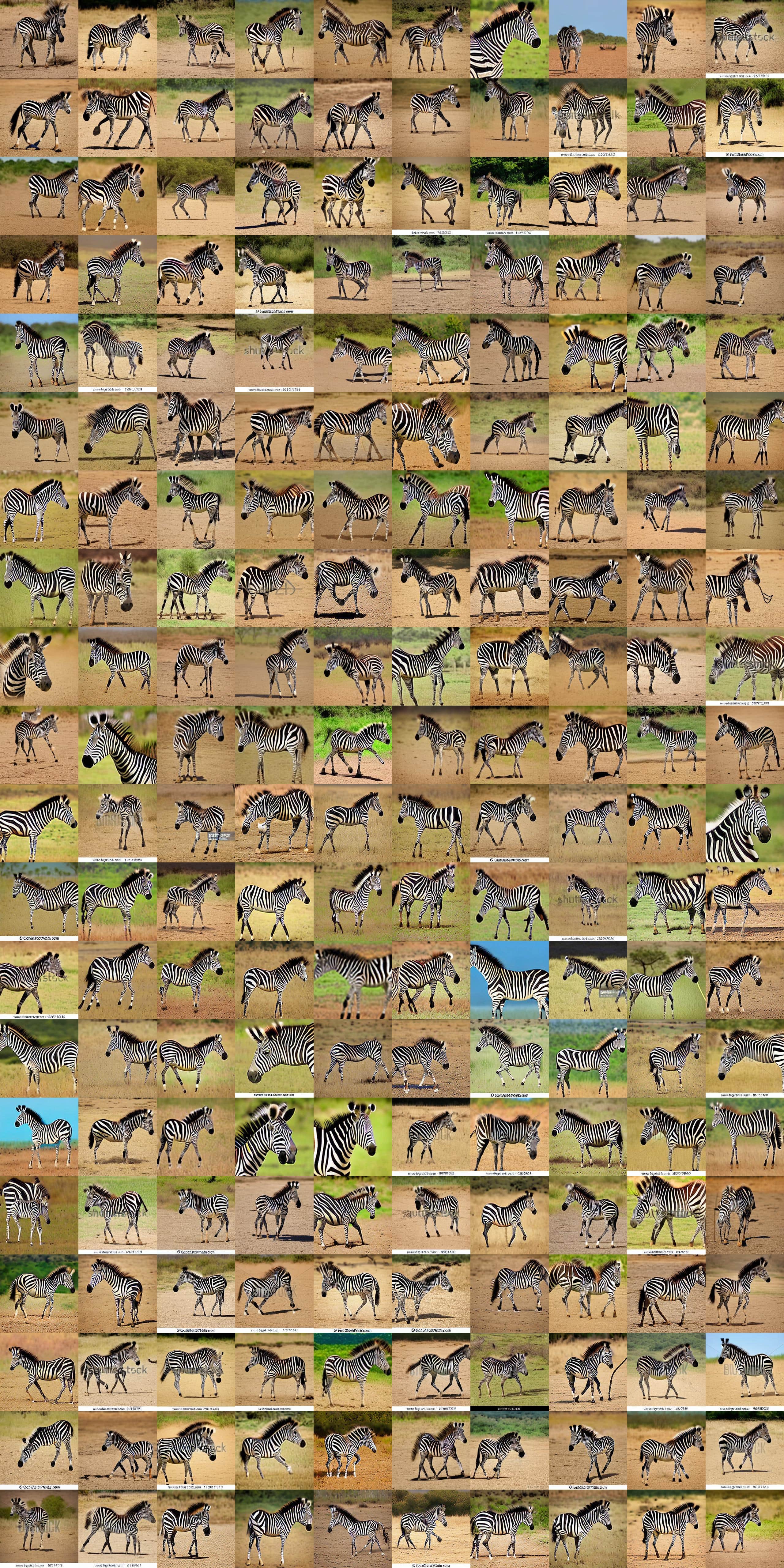}
\caption{\textbf{LDM} generated images using 4 different prompts, describing a zebra that is walking on a rope.}
\label{supp:text-200-walk}
\end{figure*}

In Figures \ref{supp:lego-200-frozen} to \ref{supp:text-200-walk}, we show all images for the concepts we used in our User + LLM study.

\section{Complete Ablation Results}
\label{supp:ablation}
\begin{figure*}
\centering
    \includegraphics[width=\linewidth]{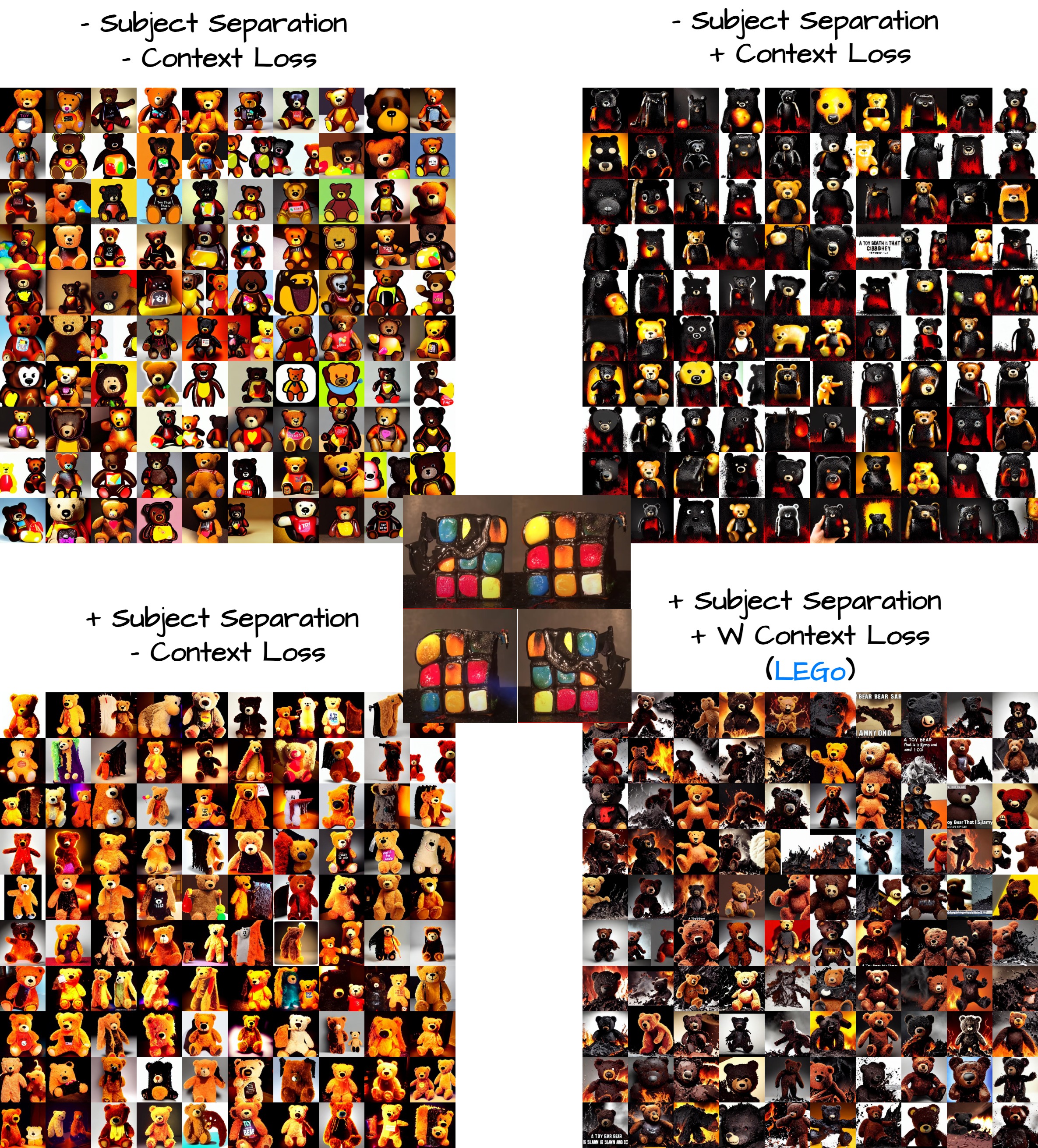}
\caption{\label{supp: single-melt} Ablation results for \textit{burnt and melted} concept from single subject example images applied to the subject ``toy bear''. Please zoom in to see details.}
\end{figure*}

\begin{figure*}
\centering
    \includegraphics[width=\linewidth]{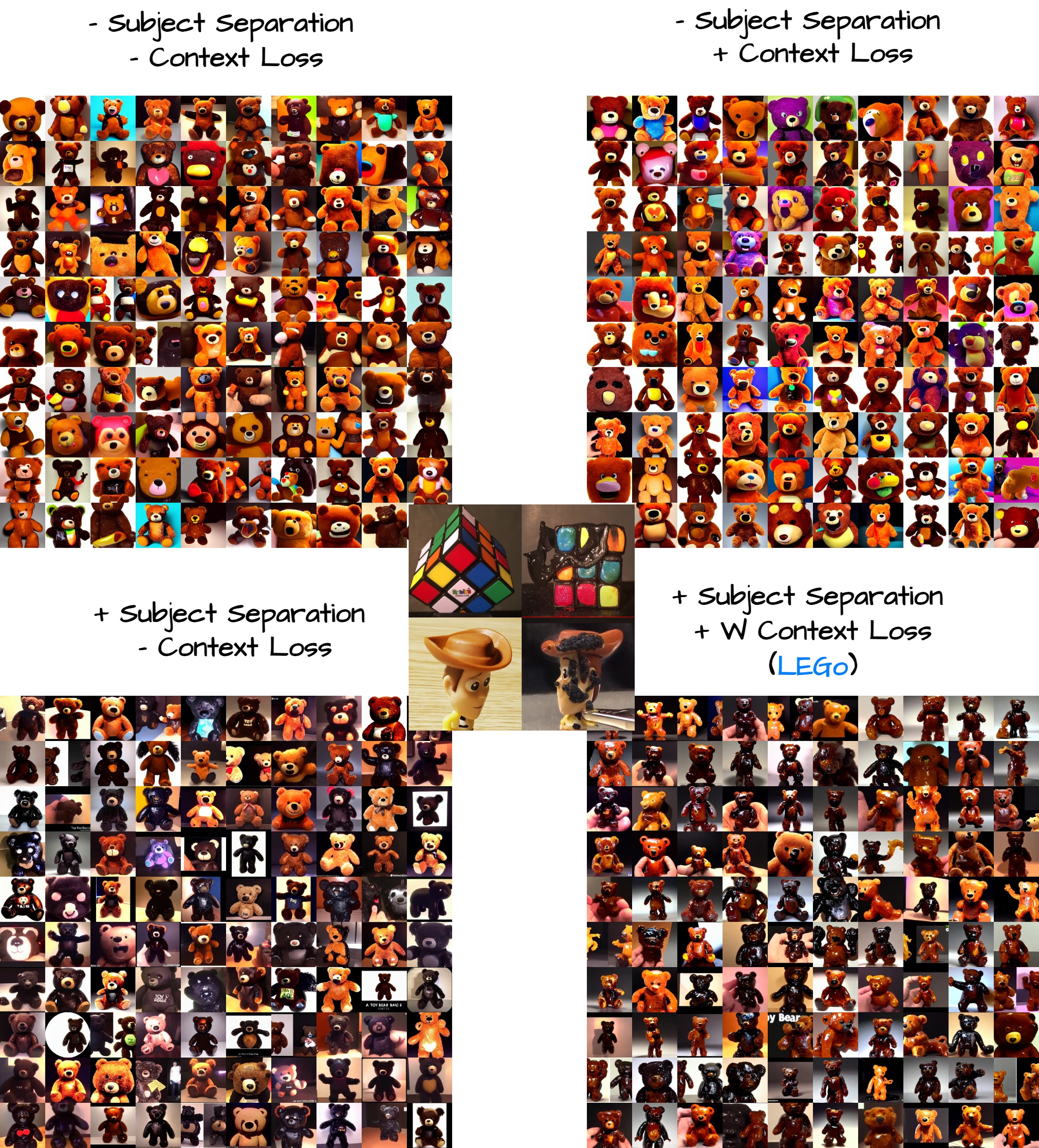}
\caption{\label{supp: SMULTI-MELT} Ablation results for \textit{burnt and melted} concept from multi subject example images applied to the subject ``toy bear''. Please zoom in to see details.}
\end{figure*}

\begin{figure*}
\centering
    \includegraphics[width=\linewidth]{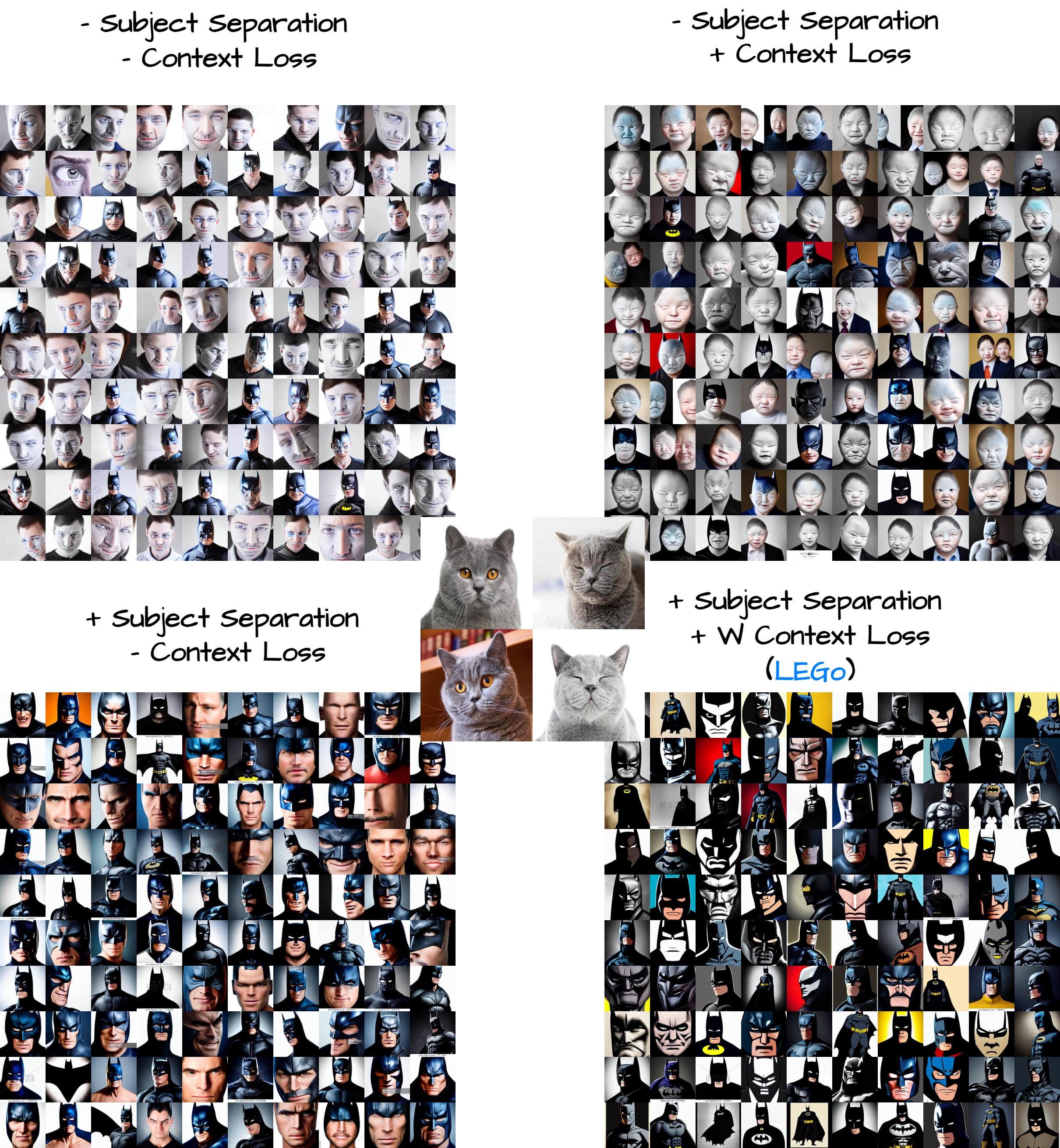}
\caption{\label{supp:ablation-eyes} Ablation results for \textit{closed eyes} concept applied to the subject ``batman''. Please zoom in to see details.}
\end{figure*}

\begin{figure*}
\centering
    \includegraphics[width=\linewidth]{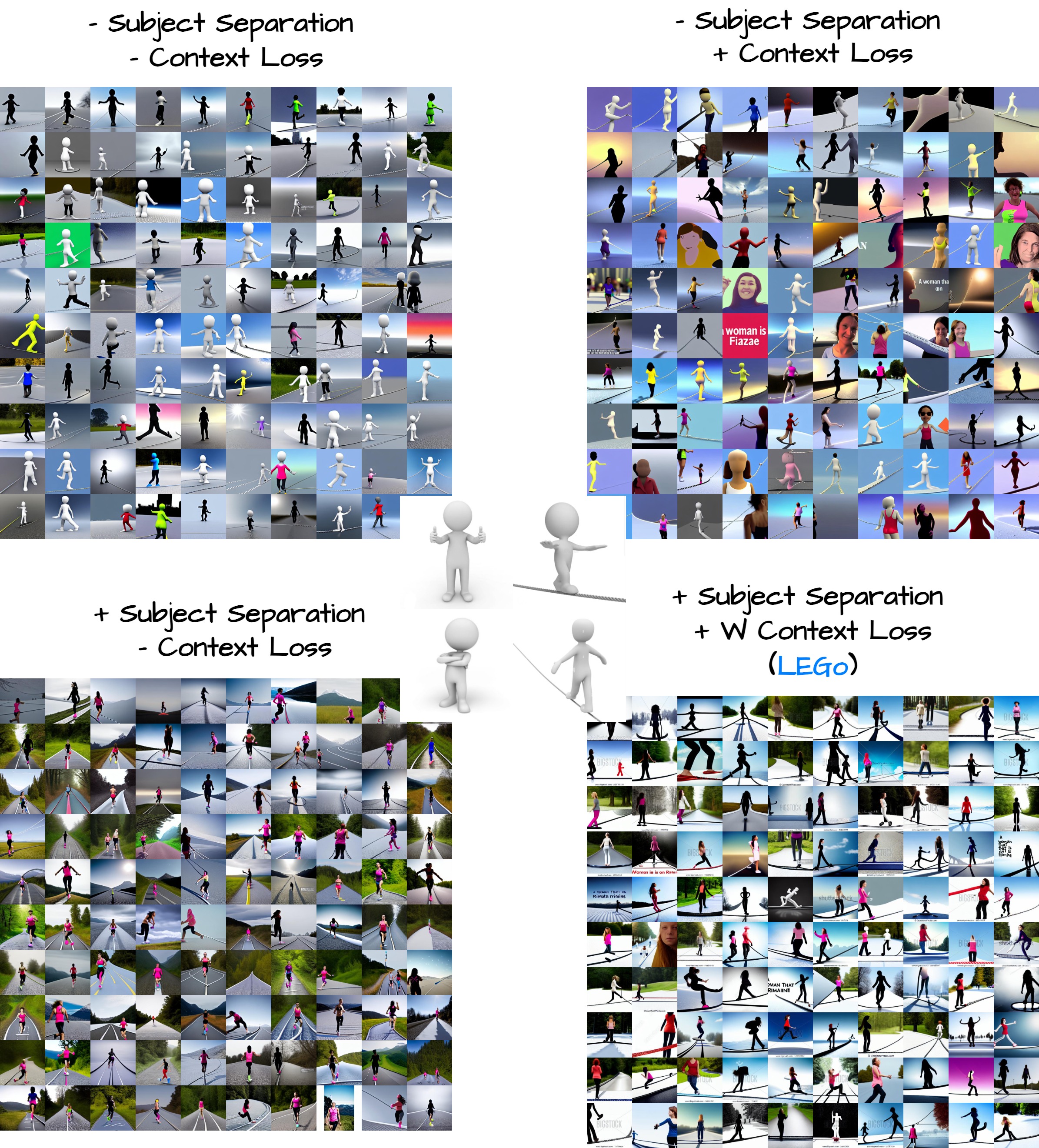}
\caption{\label{supp:ablation-walk} Ablation results for \textit{walking on rope} concept applied to the subject ``a woman''. Please zoom in to see details.}
\end{figure*}

In Section \ref{sec:ablation} we showed a few ablation results for the concept of \textit{burnt and melted}. In Figures \ref{supp: single-melt} and \ref{supp: SMULTI-MELT}, we show the full ablation results for the single and multi subject \textit{burnt and melted} concept. In the single subject experiment, we used $\mathcal{P}$ words that were more focused on \textbf{burnt} and \textbf{destroyed} whereas in the multi-subject experiment we used words focused on \textbf{melted} and \textbf{liquefy}. The effect is clear in Lego's inversion results based on the choice of positive word selection. We encourage the user to experiment with the words that they would like to prominently see in the result concept. Figures \ref{supp:ablation-eyes} and \ref{supp:ablation-walk} show the ablation results for learning the concepts of \textit{closed eyes} and \textit{walking on rope} and applying them to ``batman'' and ``a zebra'' respectively.
\\
\par\noindent The ablation for the three concepts clearly show the effect of \textbf{Subject Separation} and \textbf{Context Loss} on the concept's representation. In the single subject melting examples, we can see Rubik's cube's features in the toy bear and in the multi-subject scenario, both the features of the Rubik's cubes and Toy Story's Woody can be seen in the toy bears (notice Woody's eyes in some of the bears). Furthermore, in the \textit{closed eyes} example, the cat's features can be found in what is mean to represent Batman's face. In the \textit{walking on rope} example, the figure from the example images is present instead of ``a woman''. The effect of context loss can be seen by comparing Lego generated concepts with concepts generated by performing subject separation and not context loss. Not performing context loss leaves most images unfaithful in representing the concept.

\end{document}